\documentclass{article}

\usepackage{arxiv}

\usepackage[utf8]{inputenc} 
\usepackage[T1]{fontenc}    
\usepackage{hyperref}       
\usepackage{url}            
\usepackage{booktabs}       
\usepackage{amsfonts}       
\usepackage{nicefrac}       
\usepackage{microtype}      
\usepackage{lipsum}		
\usepackage{graphicx}
\usepackage{amsmath}
\usepackage{mathtools}
\usepackage{multirow}
\usepackage{subfig}
\usepackage{array}
\usepackage{float}
\usepackage{makecell}
\usepackage[mathscr]{eucal}
\usepackage[left]{lineno}
\title{Physics Informed Extreme Learning Machine (PIELM)-- A rapid method
	for the numerical solution of partial differential equations}

\author{
  Vikas~Dwivedi\thanks{Use footnote for providing further
    information about author (webpage, alternative
    address)---\emph{not} for acknowledging funding agencies.} \\
  Department of Mechanical Engineering \\
  Indian Institute of Technology, Madras\\
  Chennai-600036, India \\
  \texttt{me15d080@smail.iitm.ac.in} \\
   \And
 Balaji~Srinivasan \\
  Department of Mechanical Engineering\\
  Indian Institute of Technology, Madras\\
  Chennai-600036, India \\
  \texttt{sbalaji@iitm.ac.in} \\
}

\begin{document}
\maketitle

\begin{abstract}
There has been rapid progress recently on the application of deep
networks to solution of partial differential equations, collectively
labelled as Physics Informed Neural Networks (PINNs). In this paper,
we develop Physics Informed Extreme Learning Machine (PIELM), a rapid
version of PINNs which can be applied to stationary and time dependent
linear partial differential equations. We demonstrate that PIELM matches
or exceeds the accuracy of PINNs on a range of problems. We also discuss
the limitations of neural network based approaches, including our
PIELM, in the solution of PDEs on large domains and suggest an extension,
a distributed version of our algorithm -{}- DPIELM. We show that DPIELM
produces excellent results comparable to conventional numerical techniques
in the solution of time-dependent problems. Collectively, this work
contributes towards making the use of neural networks in the solution
of partial differential equations in complex domains as a competitive
alternative to conventional discretization techniques.

\end{abstract}


\section{Introduction}
Partial differential equations (PDEs) are extensively used in the
mathematical modelling of various problems in physics, engineering
and finance. In practical situations, these equations typically lack
analytical solutions and are solved numerically. In current practice,
most numerical approaches to solve PDEs like finite element method
(FEM), finite difference method (FDM) and finite volume method (FVM)
are mesh based. A typical implementation of a mesh based approach
involves three steps: (1) Grid generation, (2) Discretization of governing
equation and (3) Solution of the discretized equations with some iterative
method.

However, there are limitations to these approaches. Some of the limitations
of these methods are as follows:
\begin{enumerate}
	\item They cannot be used to solve PDEs in complex computational domains
	because grid generation (step 1) itself becomes infeasible. 
	\item The process of discretization (step 2) creates a discrepancy between
	the mathematical nature of actual PDE and its approximate difference
	equation \cite{CFD_BOOK}. Sometimes this can lead to quite serious
	problems \cite{QUIRK}. 
\end{enumerate} 

One of the options to fix these issues is to use neural networks.
In this approach, the data set consists of some randomly selected
points in the domain and on the boundary. The governing equations
and the boundary conditions are fitted using neural network. There
are two main motivations for this approach. First, being universal
approximators, neural networks can potentially represent any PDE.
So this avoids the discretization step and thus discretization based
physics errors too. Second, it is meshfree and therefore complex geometries
can be easily handled \cite{BERG ET AL}.
Initial work in this direction can be credited to Lagaris et al. \cite{LAGARIS_98,LAGARIS_2K}.
Firstly, they solved the initial boundary value problem using neural
networks and later they extended their work to handle irregular boundaries.
Since then, a lot of work has been done in this field \cite{MILLIGEN ET AL,McFALL ET AL,KUMAR ET AL,MALL ET AL,SIRIGNANO DGM,RAISSI KARNIADAKIS,RAISSI ET AL 2K18,RAISSI PINN}.
In particular, we refer to the physics-informed neural networks (PINN)
approach by Raissi and Karniadakis \cite{RAISSI KARNIADAKIS}
and Raissi et. al \cite{RAISSI ET AL 2K18,RAISSI PINN}. This
approach has produced promising results for a series of benchmark
nonlinear problems.

Recently, Berg et al. \cite{BERG ET AL} have developed a PINN
based method to solve PDEs on complex domains and produced several
results. However, in spite of various advantages of using deep networks
for solving PDEs, PINNs have several problems \cite{RAISSI PINN}.
Firstly, there is no theoretical basis to know the size of neural
network architecture and the amount of data needed. Then, there is
no guarantee that the algorithm will not hit upon a local minima.
Finally, their learning time is slower than the traditional numerical
methods making them very expensive for practical problems.

We show that some of the problems mentioned above can be easily handled
by using an alternative network called the extreme learning machine
(ELM). The basic ELM was proposed by Huang et al. \cite{HUANG ELM}
for a single hidden layer feed forward networks (SLFNs) and later
it was extended to generalized SLFNs. The essence of ELM is that the
weights of the hidden layer of SLFNs need not to be learnt. It is
much faster than the traditional gradient based optimization methods
alleviating the learning time problem. Previously, ELMs have been
used in approximating functions \cite{BALASUNDARAM} and solving
ordinary differential equations (ODEs) and stationary PDEs \cite{YANG ET. AL,SUN ET. AL}
using Legendre and Bernstein polynomial basis functions respectively.

In this paper, we propose a new machine learning algorithm to solve
stationary and time dependent PDEs in complex geometries. We have
named it physics informed extreme learning machine (PIELM) because
it is a combination of two algorithms namely ELM and PINN. Theoretically,
there is no question over the employment of ELM as a PDE solver because
it is a universal approximator \cite{HUANG UNIV APP} and therefore
it can approximate any PDE. We have made our ELM ``physics informed''
by incorporating the information about the physics of PDE as the cost
function. In addition to this, we have also proposed an extension
to original PIELM called distributed PIELM that enhances the representation
power of PIELM without adding any extra hidden layers. We demonstrate
that both PIELM and DPIELM exibit superior performance on a range
of stationary and time-dependent problems in comparison to existing
methods. 

This paper proceeds as follows. We give a brief review of PINN and
ELM in Section \ref{sec:2}. The proposed PIELM is described in Section
\ref{sec:3}.In Section \ref{sec:4}, we evaluate the performance
of PIELM in solving various stationary and time-dependent PDEs. To
our knowledge, this is the first application of an ELM based algorithm
to solve a 2D unsteady PDE. In Section \ref{sec:5}, we discuss the
limitation of PIELM to represent discontinuous functions and the functions
with sharp gradient. We also illustrate a test case where even a deep
PINN fails to represent a complicated function. In Section \ref{sec:6},
DPIELM, the distributed version of PIELM for enhanced representation
is described. In Section \ref{sec:7}, the results of implementation
of DPIELM algorithm in test cases involving representation of functions
with sharp gradients have been presented. We have also shown that
DPIELM outperforms the deep PINN in representing the complicated function
described earlier. Finally, conclusion and future work are given in
Section \ref{sec:8}.

\section{\label{sec:2}Brief review of ELM and PINN}

The PIELM is combination of two learning algorithms: ELM and PINN.
In this section, we review these two algorithms in brief. 

\subsection{Extreme learning machine}

Traditional gradient-based learning algorithms \cite{HINTONT }
have prohibitively slow learning speed and they suffer from various
problems like improper learning rate, local minima etc. Huang et al.
\cite{HUANG ELM} originally proposed a novel learning algorithm
called ELM to fix these issues. ELM is extremely fast and mostly it
shows better generalization performance than gradient-based learning
approaches like back-propagation. A typical implementation of the
algorithm involves the following steps:
\begin{enumerate}
	\item Select a shallow neural network.
	\item Fix the hidden layer weights and biases randomly. These parameters
	will not be learned and therefore no iterative optimization is required
	for them.
	\item Apply a nonlinear transformation to the input data set. This gives
	the input to the final layer. 
	\item Take the linear combination of all the inputs of the final layer.
	This is the output of ELM. 
	\item Learn the output layer weights using the least squares method.
\end{enumerate}
\begin{figure}[H]
	\begin{centering}
		\includegraphics[scale=0.4]{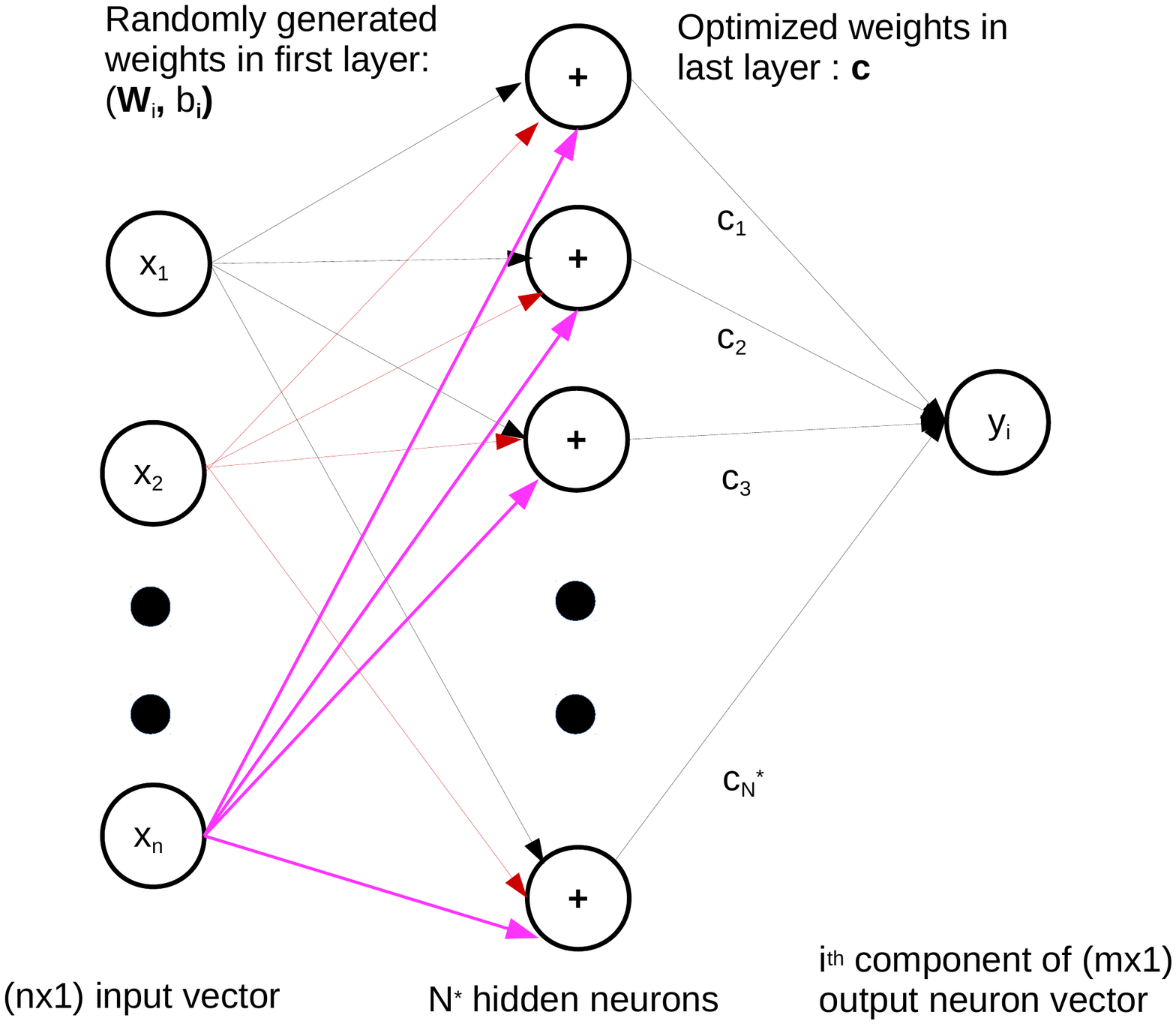}
		\par\end{centering}
	\caption{\label{fig:2.1.1_basic_ELM}Basic structure of ELM}
\end{figure}

\subsubsection*{Mathematical formulation }

Consider the basic ELM shown in Fig (\ref{fig:2.1.1_basic_ELM}).
It is a single layer feed forward neural network with $N^{*}$ neurons
in the hidden layer. Input is a vector of size $n$ and output is
the $i^{th}$ component of the output vector of size $m$. We denote
the non-linear activation by $\varphi$ and the weights and biases
of $j^{th}$ node of hidden layer by $\overrightarrow{a}_{j}^{(i)}$
and $b_{j}^{(i)}$respectively. The output of the hidden node $j$
is $\varphi(\overrightarrow{x};\overrightarrow{a}_{j}^{(i)},b_{j}^{(i)})$.
For a given data set $\{(x_{k},y_{k})\}_{k=1}^{N}\subset\Re^{n}\text{x}\Re^{m}$
with $N$ distinct samples, the ELM output is given by 
\begin{equation}
\overrightarrow{y}_{ELM}^{(i)}=\boldsymbol{H}^{(i)}(\overrightarrow{x})\overrightarrow{c}\label{eq:2.1_elm_out}
\end{equation}
where,
\[
\boldsymbol{H}^{(i)}=[\overrightarrow{h}^{(i)}(\overrightarrow{x}_{1}),\overrightarrow{h}^{(i)}(\overrightarrow{x}_{2}),...,\overrightarrow{h}^{(i)}(\overrightarrow{x}_{N})]^{T},
\]
\[
\overrightarrow{h}^{(i)}(\overrightarrow{x}_{k})=[\varphi(\overrightarrow{x}_{k};\overrightarrow{a}_{1}^{(i)},b_{1}^{(i)}),\varphi(\overrightarrow{x};\overrightarrow{a}_{2}^{(i)},b_{2}^{(i)}),...,\varphi(\overrightarrow{x};\overrightarrow{a}_{N^{*}}^{(i)},b_{N^{*}}^{(i)})],
\]
and $\overrightarrow{c}=[c_{1},c_{2},...,c_{N^{*}}]^{T}$ is vector
of output layer weights.

On writing the Eq. (\ref{eq:2.1_elm_out}) for all the $m$ components,
the resulting ELM is given by 
\begin{equation}
\boldsymbol{H}^{(i)}\overrightarrow{c}=\overrightarrow{y}^{(i)}-\overrightarrow{\xi}^{(i)},\;i=1,2,...m,
\end{equation}
where $\overrightarrow{\xi}$ is the training error vector. The ELM
tends to reach the smallest training error together with the smallest
norm of the output weights. Mathematically saying, the loss function
to be minimized for the ELM is given by
\begin{equation}
J=\frac{1}{2}||\overrightarrow{c}||^{2}+\frac{1}{2N}\lambda\sum_{i=1}^{m}\overrightarrow{\xi}^{(i)^{T}}\overrightarrow{\xi}^{(i)},
\end{equation}
where $\lambda$ is the regularization parameter. The correct weights
that minimize $J$ can be calculated by solving the normal equations
as given below. 
\begin{equation}
\frac{\partial J}{\partial c_{k}}=0,\;k=1,2,...,N^{*}
\end{equation}

\subsection{Physics informed neural network}

Raissi et al. \cite{RAISSI PINN} proposed a data efficient PINN
for approximating solutions to general non-linear PDEs and validated
it with a series of benchmark test cases. The main feature of the
PINN is the inclusion of the prior knowledge of physics in the learning
algorithm as cost function. As a result, the algorithm imposes penalty
for any non-physical solution and quickly directs it towards the correct
solution. This physics informed approach enhances the information
content of the data. As a result, the algorithm has good generalization
property even in the small data set regime. 

\subsubsection*{Mathematical formulation}

Consider a PDE of the following form

\begin{equation}
\frac{\partial}{\partial t}u(\overrightarrow{x},t)+\mathcal{\mathscr{N}}u(\overrightarrow{x},t)=R(\overrightarrow{x},t),\;(\overrightarrow{x},t)\epsilon\varOmega\text{x}[0,T],\label{eq:gen_PDE}
\end{equation}
\begin{equation}
u(\overrightarrow{x},t)=B(\overrightarrow{x},t),\;(\overrightarrow{x},t)\epsilon\partial\varOmega\text{x}[0,T],\label{eq:gen_BC}
\end{equation}
\begin{equation}
u(\overrightarrow{x},0)=F(\overrightarrow{x}),\;\overrightarrow{x}\epsilon\varOmega,\label{eq:gen_IC}
\end{equation}
where $\mathcal{\mathscr{N}}$ may be a linear or nonlinear differential
operator and $\partial\varOmega$ is the boundary of computational
domain $\Omega$. We approximate $u(\overrightarrow{x},t)$ with the
output \textbf{$f(\overrightarrow{x},t)$} of PINN. The network architecture
may be shallow or deep depending upon the non-linearity $\mathcal{\mathscr{N}}$.
The essence of PINN lies in the definition of its loss function. In
order to make the neural network ``physics informed'', the loss
function is defined such that a penalty is imposed whenever the network
output doesn't respect the physics of the problem. If we denote the
training errors in approximating the PDE, BCs and IC by $\overrightarrow{\xi}_{f}$,
$\overrightarrow{\xi}_{bc}$ and $\overrightarrow{\xi}_{ic}$ respectively.
Then, the expressions for these errors are as follows:

\begin{equation}
\overrightarrow{\xi}_{f}=\frac{\partial\overrightarrow{f}}{\partial t}+\mathcal{\mathcal{\mathscr{N}}}\overrightarrow{f}-\overrightarrow{R},\;(\overrightarrow{x},t)\epsilon\varOmega\text{x}[0,T],
\end{equation}
\begin{equation}
\overrightarrow{\xi}_{bc}=\overrightarrow{f}-\overrightarrow{B,}\;(\overrightarrow{x},t)\epsilon\partial\varOmega\text{x}[0,T],
\end{equation}

\begin{equation}
\overrightarrow{\xi}_{ic}=\overrightarrow{f}(.,0)-\overrightarrow{F},\;\overrightarrow{x}\epsilon\varOmega.
\end{equation}
For shallow networks, $\frac{\partial\overrightarrow{f}}{\partial t}$
and $\mathcal{\mathcal{\mathscr{N}}}\overrightarrow{f}$ can be determined
using hand calculations. However, for deep networks, we have to use
automatic differentiation \cite{BAYDIN}. The loss function $J$
to be minimized for a PINN is given by
\begin{equation}
J=\frac{\overrightarrow{\xi}_{f}^{T}\overrightarrow{\xi}_{f}}{2N_{f}}+\frac{\overrightarrow{\xi}_{bc}^{T}\overrightarrow{\xi}_{bc}}{2N_{bc}}+\frac{\overrightarrow{\xi}_{ic}^{T}\overrightarrow{\xi}_{ic}}{2N_{ic}},
\end{equation}
where $N_{f}$, $N_{bc}$ and $N_{ic}$ refer to number of collocation
points, boundary condition points and initial condition points respectively.
Finally, any gradient based optimization routine may be used to minimize
$J$. 

This completes the mathematical formulation of PINN. The key steps
in its implementation are as follows:
\begin{enumerate}
	\item Identify the PDE to be solved along with the initial and boundary
	conditions. 
	\item Decide the architecture of PINN.
	\item Approximate the correct solution with PINN.
	\item Find expressions for the PDE, BCs and IC in terms of PINN and its
	derivatives. 
	\item Define a loss function which penalizes for error in PDE, BCs and IC.
	\item Minimize the loss with gradient based algorithms.
\end{enumerate}

\section{\label{sec:3}Proposed PIELM}

\begin{figure}
	\begin{centering}
		\includegraphics[scale=0.4]{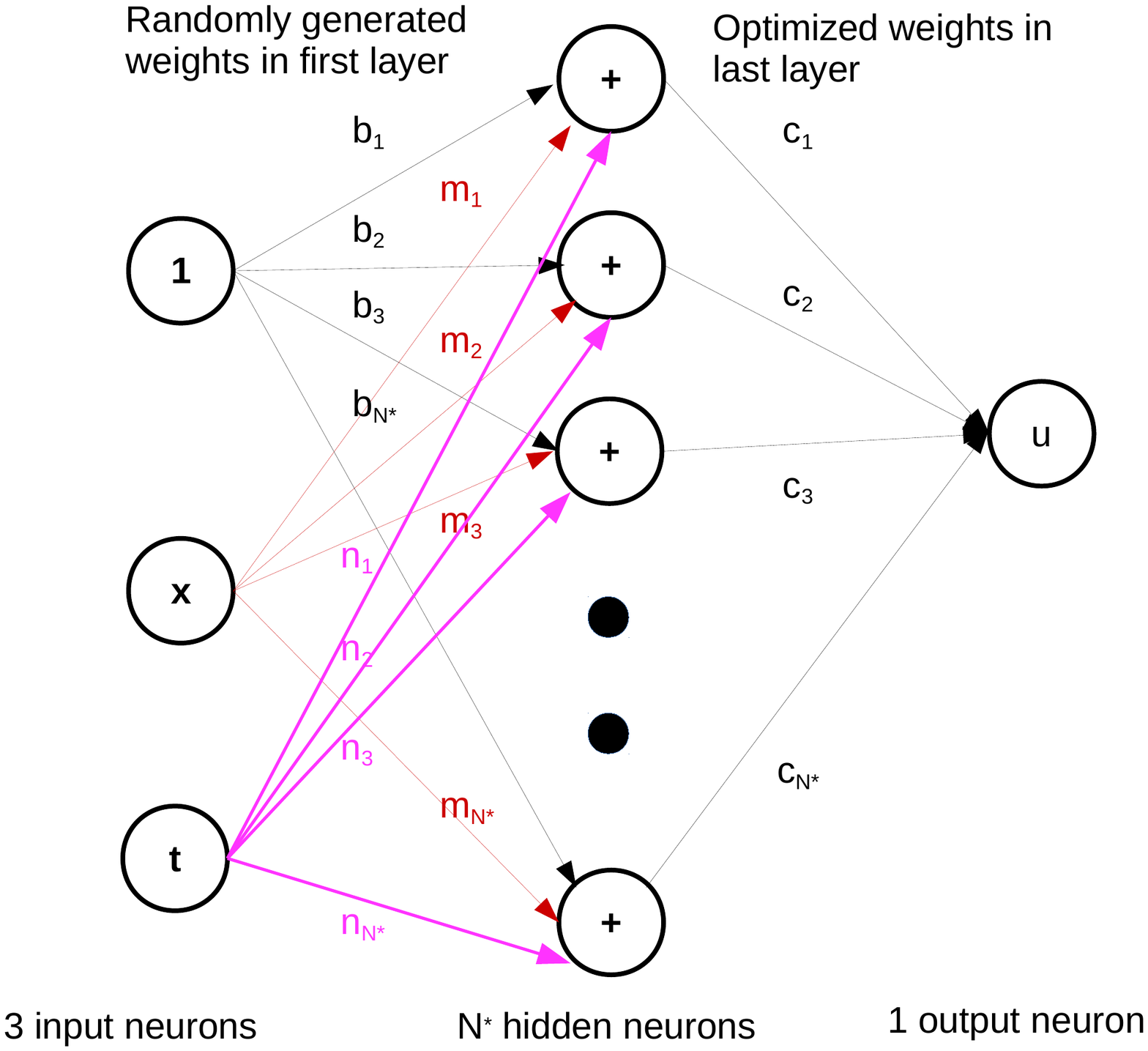}
		\par\end{centering}
	\caption{\label{fig:PiELM_2D_net}PIELM for 1D unsteady problems }
\end{figure}

Consider the following unsteady linear PDE 
\begin{equation}
\frac{\partial}{\partial t}u(\overrightarrow{x},t)+\mathcal{L}u(\overrightarrow{x},t)=R(\overrightarrow{x},t),(\overrightarrow{x},t)\epsilon\varOmega\text{x}[0,T],\label{eq:gen_PDE-1}
\end{equation}
\begin{equation}
u(\overrightarrow{x},t)=B(\overrightarrow{x},t),(\overrightarrow{x},t)\epsilon\partial\varOmega\text{x}[0,T],\label{eq:gen_BC-1}
\end{equation}
\begin{equation}
u(\overrightarrow{x},0)=F(\overrightarrow{x}),\overrightarrow{x}\epsilon\varOmega,\label{eq:gen_IC-1}
\end{equation}
where $\mathcal{L}$ is a linear differential operator and $\partial\varOmega$
is the boundary of computational domain $\Omega$. We approximate
$u(\overrightarrow{x},t)$ with the output \textbf{$f(\overrightarrow{x},t)$}
of PIELM. For simplicity, we consider the 1D unsteady version of Eqns
(\ref{eq:gen_PDE-1}, \ref{eq:gen_BC-1}, \ref{eq:gen_IC-1}). The
extension to higher dimensional problems is straightforward. The PIELM
for 1D unsteady problem is schematically shown in Fig (\ref{fig:PiELM_2D_net}).
The number of neurons in the hidden layers is $N^{*}$. If we define
$\overrightarrow{\chi}=[x,t,1]^{T}$, $\overrightarrow{m}=[m_{1},m_{2},...,m_{N^{*}}]^{T}$,
$\overrightarrow{n}=[n_{1},n_{2},...,n_{N^{*}}]^{T}$, $\overrightarrow{b}=[b_{1},b_{2},...,b_{N^{*}}]^{T}$
and $\overrightarrow{c}=[c_{1},c_{2},...,c_{N^{*}}]^{T}$ then, the
output of the $k^{th}$ hidden neuron is 
\[
h_{k}=\varphi(z_{k}),
\]
where $z_{k}=[m_{k},n_{k},b_{k}]\overrightarrow{\chi}$ and $\varphi=tanh$
is the nonlinear activation function. The PIELM output is given by
\begin{equation}
f(\overrightarrow{\chi})=\overrightarrow{h}\overrightarrow{c}.
\end{equation}
Similarly, the formulae for $\frac{\partial^{p}f}{\partial x^{p}}$
and $\frac{\partial f}{\partial t}$ are given by 
\begin{equation}
\frac{\partial^{p}f_{k}}{\partial x^{p}}=m_{k}^{p}\frac{\partial^{p}\varphi}{\partial z^{p}},
\end{equation}
\begin{equation}
\frac{\partial f_{k}}{\partial t}=n_{k}\frac{\partial\varphi}{\partial z}.
\end{equation}
We denote the training errors in approximating the PDE, BCs and IC
by $\overrightarrow{\xi}_{f}$, $\overrightarrow{\xi}_{bc}$ and $\overrightarrow{\xi}_{ic}$
respectively. The expressions for these errors are as follows:

\begin{equation}
\overrightarrow{\xi}_{f}=\frac{\partial\overrightarrow{f}}{\partial t}+\mathcal{L}\overrightarrow{f}-\overrightarrow{R},\;(\overrightarrow{x},t)\epsilon\varOmega\text{x}[0,T],
\end{equation}
\begin{equation}
\overrightarrow{\xi}_{bc}=\overrightarrow{f}-\overrightarrow{B,}\;(\overrightarrow{x},t)\epsilon\partial\varOmega\text{x}[0,T],
\end{equation}

\begin{equation}
\overrightarrow{\xi}_{ic}=\overrightarrow{f}(.,0)-\overrightarrow{F},\;\overrightarrow{x}\epsilon\varOmega.
\end{equation}
Next, we put a hard constraint on $\overrightarrow{c}$ to solve the
PDE exactly with zero error by setting 

\begin{equation}
\overrightarrow{\xi}_{f}=\overrightarrow{0},\label{eq:err_f=00003D0}
\end{equation}

\begin{equation}
\overrightarrow{\xi}_{bc}=\overrightarrow{0},\label{eq:err_bc=00003D0}
\end{equation}

\begin{equation}
\overrightarrow{\xi}_{ic}=\overrightarrow{0}.\label{eq:err_ic=00003D0}
\end{equation}
Eqns (\ref{eq:err_f=00003D0} to \ref{eq:err_ic=00003D0}) lead to
the a system of linear equations which can be represented as
\begin{equation}
\boldsymbol{H}\overrightarrow{c}=\overrightarrow{K}.\label{eq:Hc=00003DK}
\end{equation}
The form of $\boldsymbol{H}$ and $\overrightarrow{K}$ depends on
the $\mathcal{L}$, $B$ and $F$ i.e. on the type of PDE, boundary
condition and initial condition. In order to find $\overrightarrow{c}$,
Moore--Penrose generalized inverse \cite{PSEUDO INVERSE BOOK}
(also called pseudo-inverse) should be used as it works well for singular
and non square $\boldsymbol{H}$ too. An additional advantage with
this formulation is that we have a basis to guess the scale of the
PIELM architecture. When we are solving Eqns (\ref{eq:err_f=00003D0}
to \ref{eq:err_ic=00003D0}) simultaneously, we know that a unique
solution will exist when number of unknowns are equal to number of
equations which means that $N^{*}=N_{f}+N_{bc}+N_{ic}$. This gives
us an idea of the size of the hidden layer. However, in practice we
get the correct solution even with lesser number of neurons. For example,
if we supply a large number of points to approximate a linear function,
the PIELM would not require the same number of neurons for learning. 

This completes the mathematical formulation of PIELM. The key steps
in its implementation are as follows:
\begin{enumerate}
	\item Assign the input layer weights randomly.
	\item Depending on the PDE and the initial and boundary conditions, find
	the expressions for $\overrightarrow{\xi}_{f}$, $\overrightarrow{\xi}_{bc}$
	and $\overrightarrow{\xi}_{ic}$.
	\item Assemble the three sets of equations in the form of $\boldsymbol{H}\overrightarrow{c}=\overrightarrow{K}.$ 
	\item Output layer weight vector is given by $pinv(\boldsymbol{H})\overrightarrow{K},$
	where $pinv$ refers to pseudo-inverse.
\end{enumerate}
It is to be noted that unlike conventional ELMs , we are not solving
an optimization problem. The loss function $J$ to be minimized for
a conventional physics informed ELM would be given by
\begin{equation}
J=\frac{1}{2}||\overrightarrow{c}||^{2}+\frac{1}{2}\lambda\left(\frac{\overrightarrow{\xi}_{f}^{T}\overrightarrow{\xi}_{f}}{2N_{f}}+\frac{\overrightarrow{\xi}_{bc}^{T}\overrightarrow{\xi}_{bc}}{2N_{bc}}+\frac{\overrightarrow{\xi}_{ic}^{T}\overrightarrow{\xi}_{ic}}{2N_{ic}}\right),
\end{equation}
where $\lambda$ is a regularization parameter and $N_{f}$, $N_{bc}$
and $N_{ic}$ refer to number of collocation points, boundary condition
points and initial condition points respectively. The correct ELM
weights that minimize $J$ can be calculated by solving the normal
equations which is given below.
\begin{equation}
\frac{\partial J}{\partial c_{k}}=0,\;k=1,2,...,N^{*}
\end{equation}

Although a PIELM can be made with this minimization approach, we have
opted for the direct approach due to the following reasons:
\begin{enumerate}
	\item The direct approach is straightforward to formulate and code. It saves
	the effort of calculating loss function and setting the derivatives
	equal to zero.
	\item The learning of the minimization approach is comparatively less ``physics
	informed'' because physics is not being imposed in an exact sense.
\end{enumerate}
\section{\label{sec:4}Performance evaluation of PIELM}

\begin{figure}[H]
	\subfloat[\label{fig:PIELM-1D-steady}For 1D steady problems]{\includegraphics[scale=0.28]{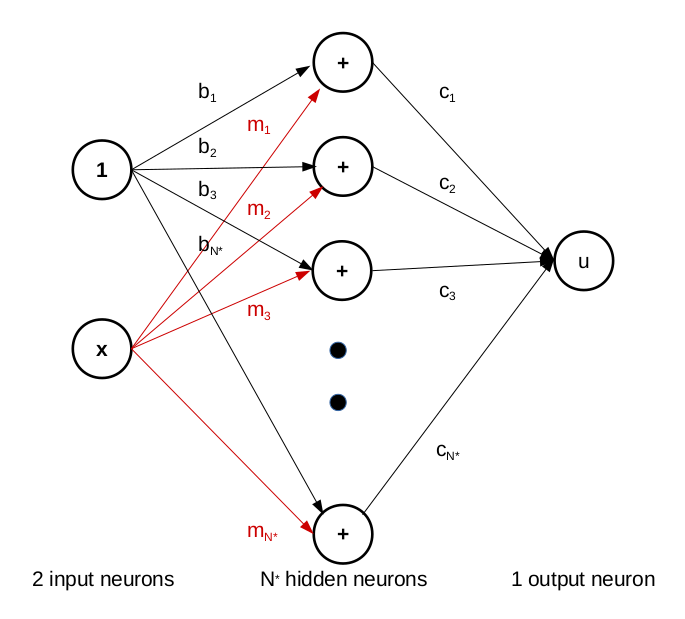}		
	}\hfill{}\subfloat[\label{fig:PIELM-2D-unsteady}For 2D unsteady problems]{\includegraphics[scale=0.33]{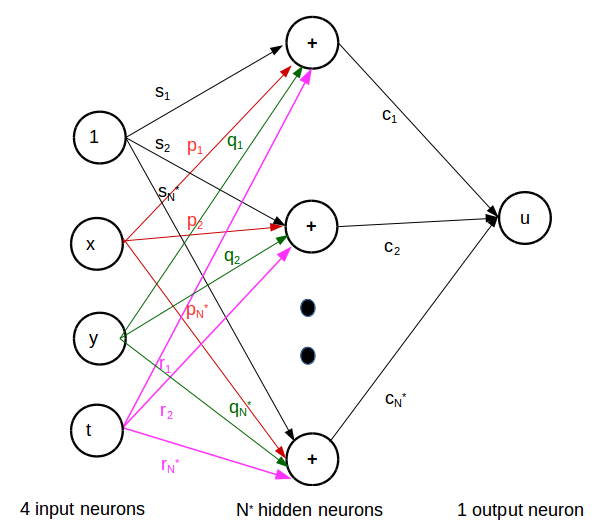}
		
	}
	
	\caption{PIELMs for steady 1D and unsteady 2D problems }
\end{figure}

\begin{table}
	\begin{centering}
		\begin{tabular}{|>{\raggedright}p{1.5cm}|>{\raggedright}p{1cm}|>{\centering}p{1.5cm}|>{\centering}p{5cm}|}
			\hline 
			Steady/
			
			Unsteady & 1D/2D & Test
			
			case 
			
			ID & Description\tabularnewline
			\hline 
			\hline 
			\multirow{6}{1.5cm}{Steady} & \multirow{3}{1cm}{1D} & TC-1 & Linear advection \tabularnewline
			\cline{3-4} 
			&  & TC-2 & Linear diffusion\tabularnewline
			\cline{3-4} 
			&  & TC-3 & Linear advection-diffusion\tabularnewline
			\cline{2-4} 
			& \multirow{3}{1cm}{2D} & TC-4 & Linear advection in a star shaped computational domain \tabularnewline
			\cline{3-4} 
			&  & TC-5 & Linear diffusion in a star shaped computational domain\tabularnewline
			\cline{3-4} 
			&  & TC-6 & Linear diffusion in a complex computational domain\tabularnewline
			\hline 
			\multirow{4}{1.5cm}{Unsteady} & \multirow{3}{1cm}{1D} & TC-7 & Linear advection \tabularnewline
			\cline{3-4} 
			&  & TC-8 & Quasi-linear advection\tabularnewline
			\cline{3-4} 
			&  & TC-9 & Linear advection-diffusion\tabularnewline
			\cline{2-4} 
			& 2D & TC-10 & Linear advection-diffusion\tabularnewline
			\hline 
		\end{tabular}
		\par\end{centering}
	\caption{\label{tab:TC_PIELM}List of test cases for PIELM.}
\end{table}
To evaluate the performance of PIELM, we rigorously test it on various
linear and quasi-linear PDEs described in Table(\ref{tab:TC_PIELM}).
TC-1, TC-2, TC-4, TC-5, TC-6 are taken from Berg et al. \cite{BERG ET AL}.
TC-8 is taken from Kopriva et al. \cite{KOPRIVA}. TC-9 and TC-10
are taken from Borker et al. \cite{FARHAT}. All the experiments
are conducted in Matlab 2017b environment running in an Intel Core
i5 2.20GHz CPU and 8GB RAM Dell laptop. The error is defined as the
difference between the PIELM prediction and the exact solution. 

\subsection{1D steady cases {[} TC-1, TC-2, TC-3 {]}}

\begin{figure}[H]
	\subfloat[\label{fig:4.1.1a_solution}PIELM solution.]{\begin{raggedright}
			\includegraphics[scale=0.65]{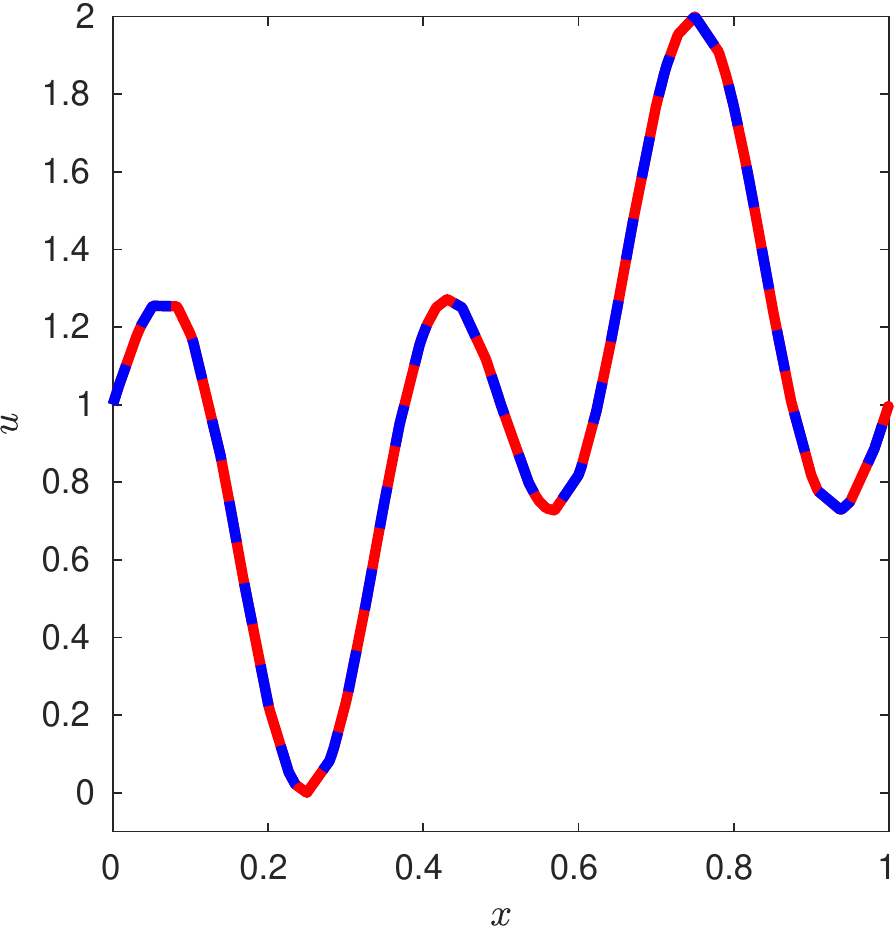}
			\par\end{raggedright}
	}\hfill{}\subfloat[\label{fig:4.1.1b_error}Error plot.]{\begin{raggedleft}
			\includegraphics[scale=0.65]{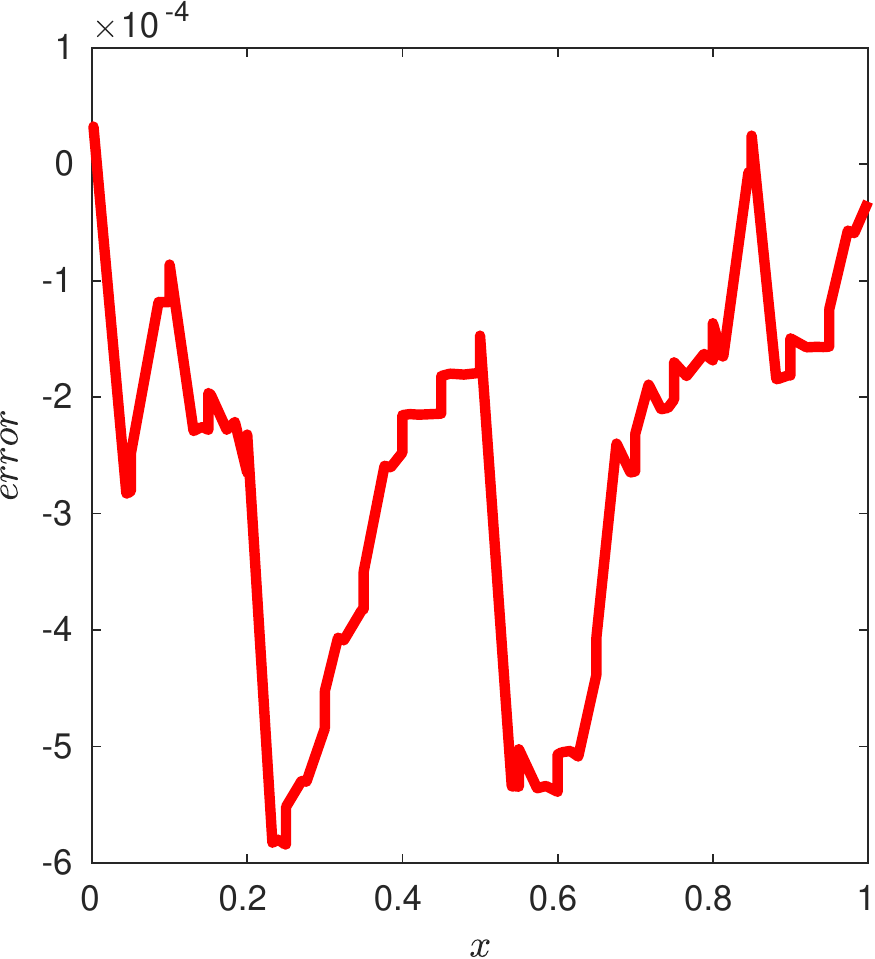}
			\par\end{raggedleft}
	}
	
	\caption{\label{fig:4.1.1_1D_steady_advection}Solution and error for steady
		1D advection. Red: PIELM, Blue: Exact.}
\end{figure}
\begin{figure}[H]
	\subfloat[\label{fig:4.1.2a}PIELM solution.]{\begin{raggedright}
			\includegraphics[scale=0.65]{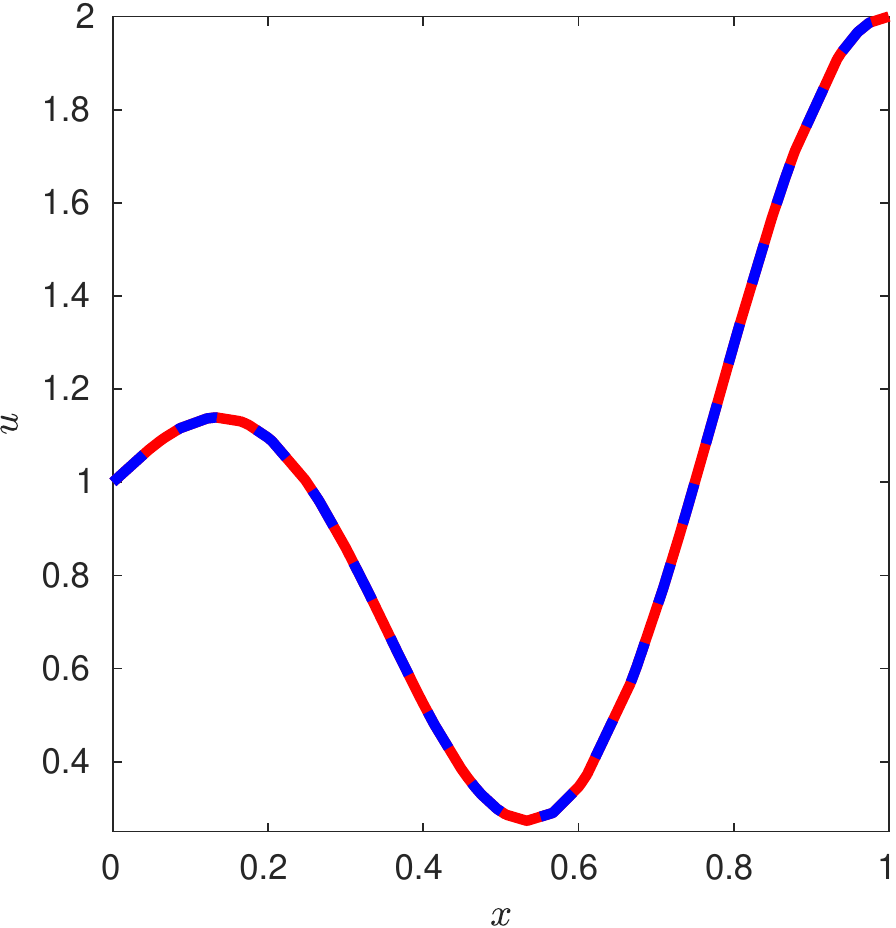}
			\par\end{raggedright}
	}\hfill{}\subfloat[\label{fig:4.1.2b}Error plot.]{\begin{raggedleft}
			\includegraphics[scale=0.65]{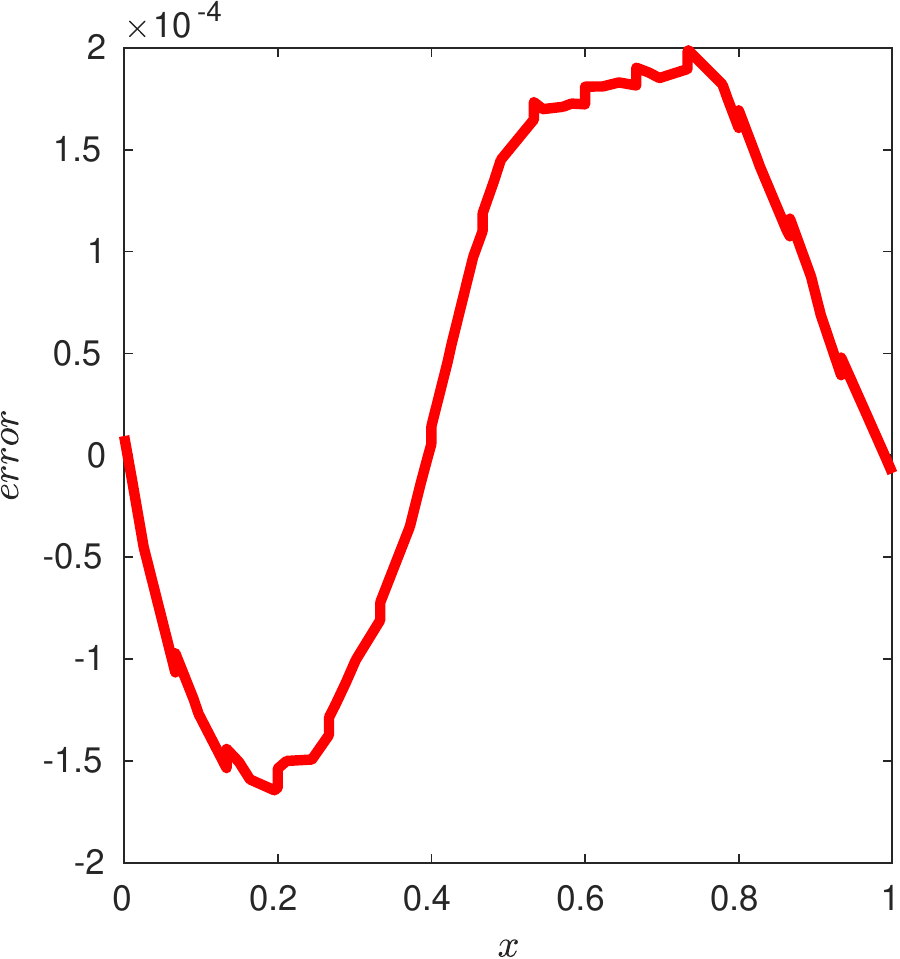}
			\par\end{raggedleft}
	}
	
	\caption{\label{fig:4.1.2_1d_diffu}Solution and error for steady 1D diffusion.
		Red: PIELM, Blue: Exact.}
\end{figure}

\begin{figure}[H]
	\subfloat[\label{fig:4.1.1a_solution-1}PIELM solution.]{\begin{raggedright}
			\includegraphics[scale=0.65]{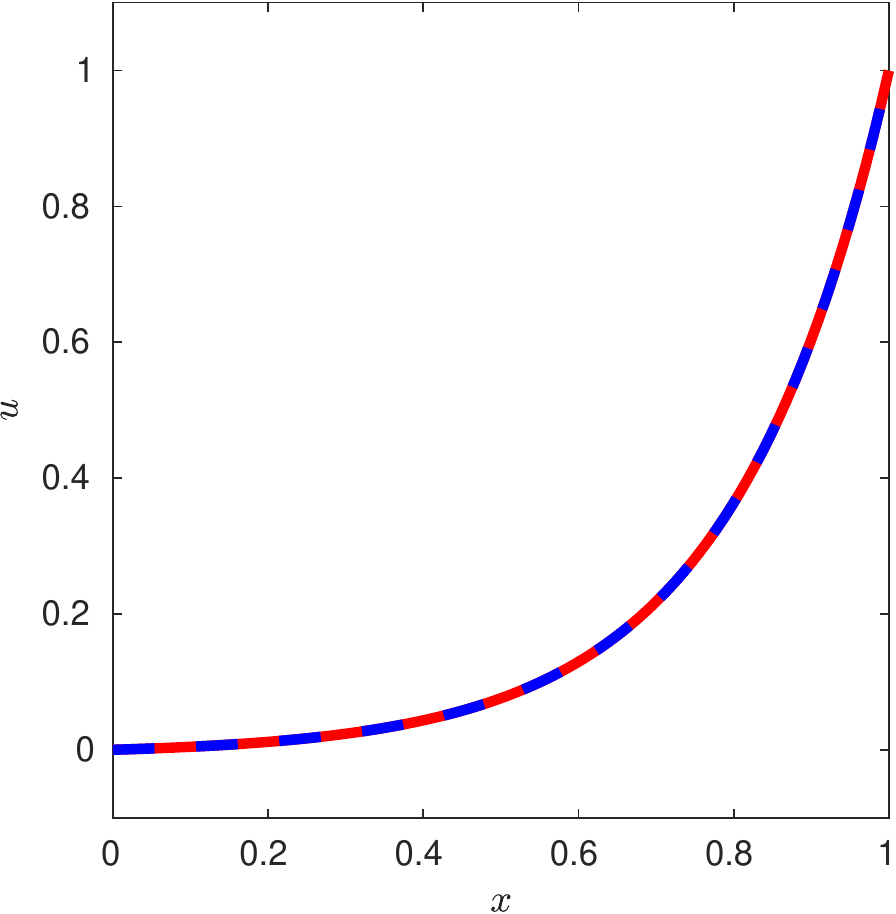}
			\par\end{raggedright}
	}\hfill{}\subfloat[\label{fig:4.1.1b_error-1}Error plot.]{\begin{raggedleft}
			\includegraphics[scale=0.65]{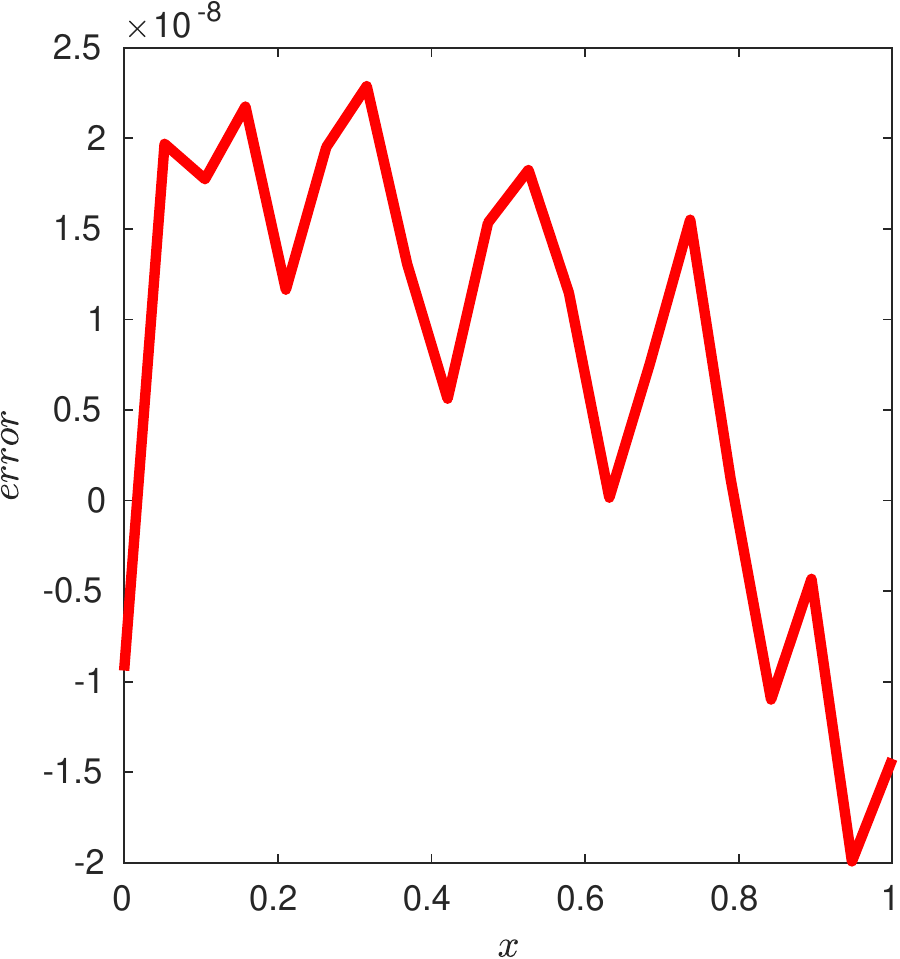}
			\par\end{raggedleft}
	}
	
	\caption{\label{fig:4.1.1_1D_SAD}Solution and error for 1D steady advection
		diffusion at $\nu=0.2$. Red: PIELM, Blue: Exact.}
\end{figure}
The 1D stationary advection, diffusion and advection-diffusion equations
are given by
\begin{equation}
u_{x}=R,0<x\leq1,
\end{equation}
\begin{equation}
u_{xx}=R,0<x\leq1,
\end{equation}
\begin{equation}
u_{x}-\nu u_{xx}=R,0<x<1
\end{equation}
respectively. The expressions for $R$ and the Dirichlet boundary
conditions for these cases are calculated by assuming the following
exact solutions
\begin{equation}
\widehat{u}=sin(2\pi x)cos(4\pi x)+1,
\end{equation}

\begin{equation}
\widehat{u}=sin(\frac{\pi x}{2})cos(2\pi x)+1,
\end{equation}
\begin{equation}
\widehat{u}=\frac{e^{\frac{x}{\nu}}-1}{e^{\frac{1}{\nu}}-1}
\end{equation}
respectively. In order to solve these equations in PIELM framework,
we have to solve Eqn(\ref{eq:err_f=00003D0} to \ref{eq:err_ic=00003D0}).
The expression for $\overrightarrow{\xi}_{f}$ depends on the linear
differential operator $\mathcal{L}$. The definitions of $\mathcal{L}$
in these cases are $\frac{\partial}{\partial x}$, $\frac{\partial^{2}}{\partial x^{2}}$
and $\frac{\partial}{\partial x}-\nu\frac{\partial^{2}}{\partial x^{2}}$
respectively. When $\mathcal{L}$ acts on $u$, the corresponding
expressions for $\overrightarrow{\xi}_{f}=\overrightarrow{0}$ may
be written as follows:
\begin{equation}
\varphi'(\boldsymbol{X_{f}}\boldsymbol{W}^{T})\text{\ensuremath{\odot}}\overrightarrow{c}\text{\ensuremath{\odot}}\overrightarrow{m}=R(\overrightarrow{x}_{f}),
\end{equation}
\begin{equation}
\varphi''(\boldsymbol{X_{f}}\boldsymbol{W}^{T})\text{\ensuremath{\odot}}\overrightarrow{c}\text{\ensuremath{\odot}}\overrightarrow{m}\text{\ensuremath{\odot\overrightarrow{m}}}=R(\overrightarrow{x}_{f}),
\end{equation}
\begin{equation}
\varphi'(\boldsymbol{X_{f}}\boldsymbol{W}^{T})\text{\ensuremath{\odot}}\overrightarrow{c}\text{\ensuremath{\odot}}\overrightarrow{m}-\nu\varphi''(\boldsymbol{X_{f}}\boldsymbol{W}^{T})\text{\ensuremath{\odot}}\overrightarrow{c}\text{\ensuremath{\odot}}\overrightarrow{m}\text{\ensuremath{\odot}}\overrightarrow{m}=\overrightarrow{0}.
\end{equation}
where $\overrightarrow{x}_{f}$ is collocation points vector, $\overrightarrow{I}$
is bias vector, $\boldsymbol{X_{f}}=[\overrightarrow{x}_{f},\overrightarrow{I}]$
and '$\text{\ensuremath{\odot}}$' refers to Hadamardt product. Referring
to Fig (\ref{fig:PIELM-1D-steady}), $\boldsymbol{W}=[\overrightarrow{m},\overrightarrow{b}]$.
Similarly, expression for $\overrightarrow{\xi}_{bc}=\overrightarrow{0}$
is given by 
\begin{equation}
\varphi(\boldsymbol{X_{bc}}\boldsymbol{W}^{T})\overrightarrow{c}=B(\overrightarrow{x}_{bc})
\end{equation}
where $\overrightarrow{x}_{bc}$ is boundary points vector, $\boldsymbol{X_{bc}}=[\overrightarrow{x}_{bc},\overrightarrow{I}]$
and $B$ is the boundary condition. 

The results for these test cases are given in Fig(\ref{fig:4.1.1_1D_steady_advection}),
Fig(\ref{fig:4.1.2_1d_diffu}) and Fig (\ref{fig:4.1.1_1D_SAD}) respectively
and the summary of the experiments is given in Tab(\ref{tab:1d_steady_results}).
\begin{table}[H]
	\begin{centering}
		\begin{tabular}{|>{\centering}p{1cm}|>{\centering}p{3cm}|>{\centering}p{3cm}|}
			\hline 
			TC & $[N_{f},N_{bc},N^{*}]$ & $\mathcal{O}(Error)$\tabularnewline
			\hline 
			\hline 
			TC-1 & $[40,2,42]$ & $10^{-4}$\tabularnewline
			\hline 
			TC-2 & $[40,2,42]$ & $10^{-4}$\tabularnewline
			\hline 
			TC-3 & $[20,2,22]$ & $10^{-6}$\tabularnewline
			\hline 
		\end{tabular}
		\par\end{centering}
	\caption{\label{tab:1d_steady_results}Summary of experiments for 1D steady
		test cases.}
\end{table}

\subsubsection*{Remark }
\begin{enumerate}
	\item It should be noted that the unified deep ANN algorithm \cite{BERG ET AL}
	took 100 points to achieve an order of accuracy of $10^{-5}$ and
	$10^{-3}$ in TC-1 and TC-2 respectively. In comparison, PIELM took
	less than half of the points and still achieved an order of accuracy
	of $10^{-4}$ in both the cases.
\end{enumerate}

\subsection{2D steady cases {[} TC-4, TC-5, TC-6 {]}}

\begin{figure}[H]
	\subfloat[$\Omega_{1}$]{\begin{raggedright}
			\includegraphics[scale=0.47]{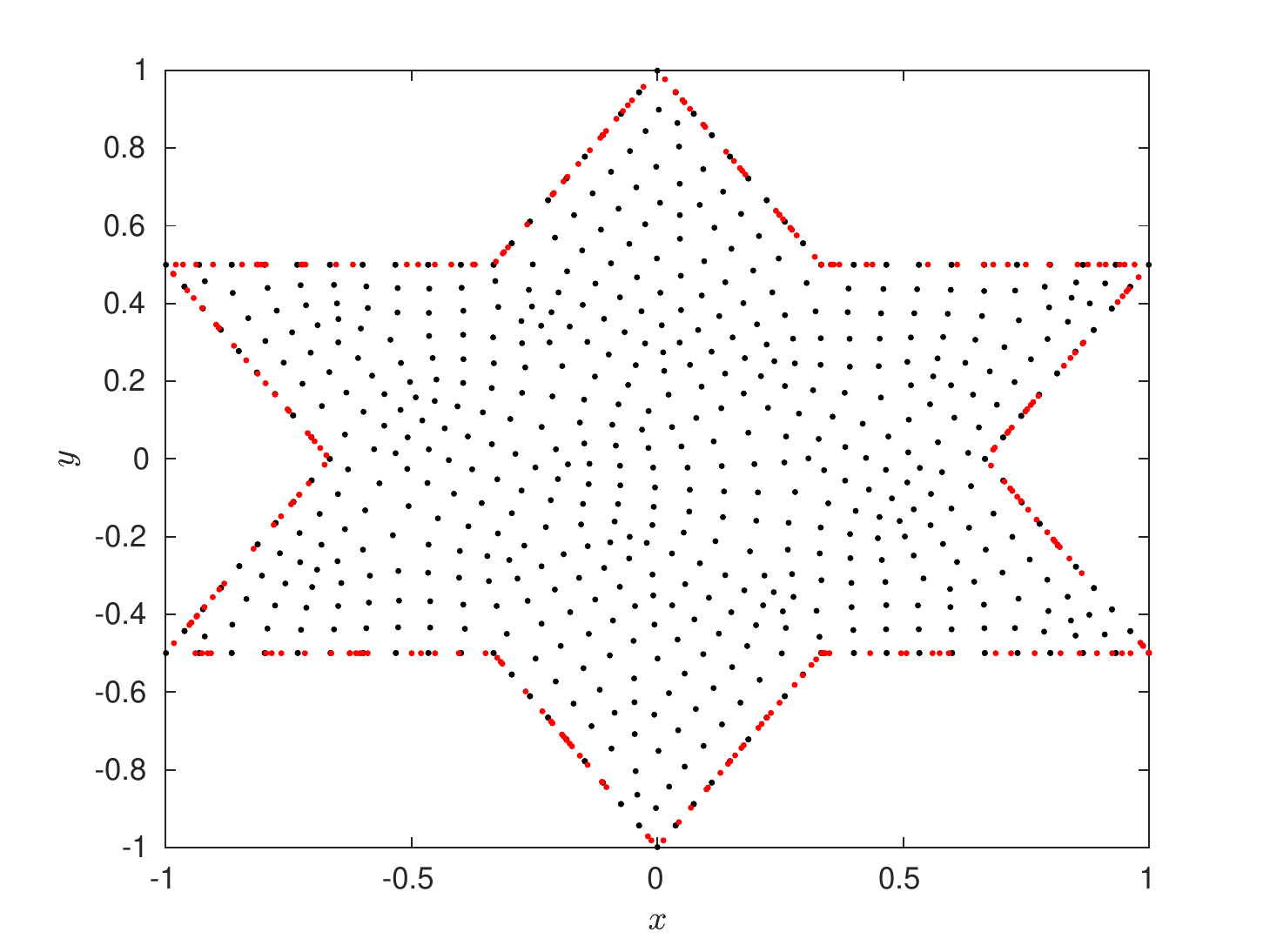}
			\par\end{raggedright}
		
	}\hfill{}\subfloat[$\Omega_{2}$]{\begin{raggedleft}
			\includegraphics[scale=0.47]{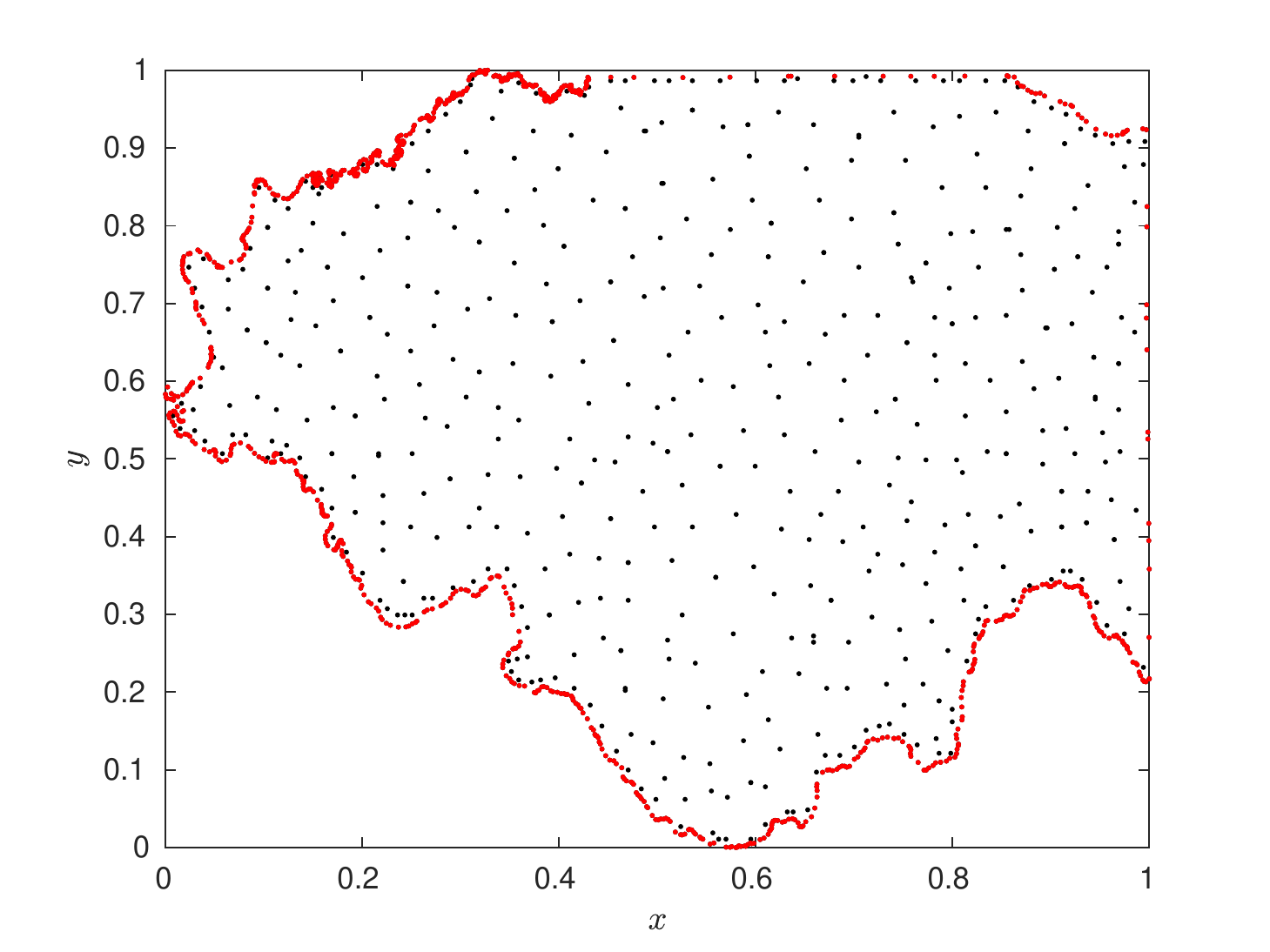}
			\par\end{raggedleft}
		
	}
	
	\caption{Computational domains for 2D steady problems}
	
\end{figure}
\begin{figure}[H]
	\subfloat[\label{fig:PIELM_sol_star_adv}PIELM solution. ]{\includegraphics[scale=0.4]{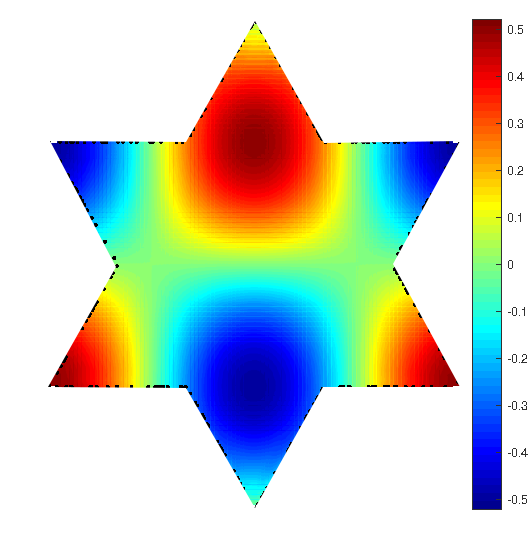}
		
	}\hfill{}\subfloat[\label{fig:PIELM_err_star_adv}Error plot.]{\includegraphics[scale=0.4]{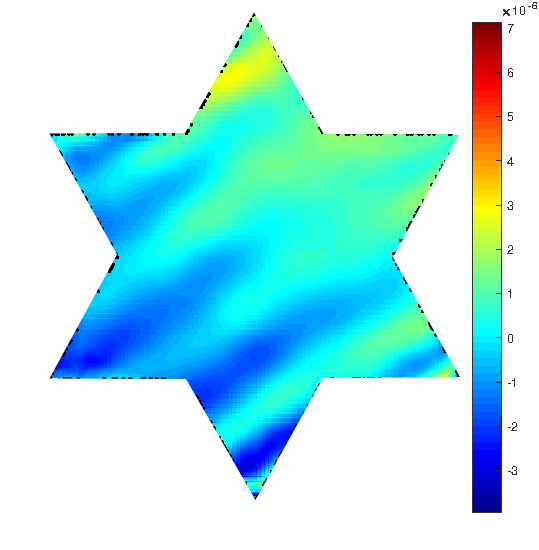}
		
	}
	
	\caption{\label{fig:8B}Solution and error for 2D steady advection equation
		on $\varOmega_{1}$.}
\end{figure}
\begin{figure}[H]
	\subfloat[\label{fig:PIELM_sol_star_diff}PIELM solution. ]{\includegraphics[scale=0.4]{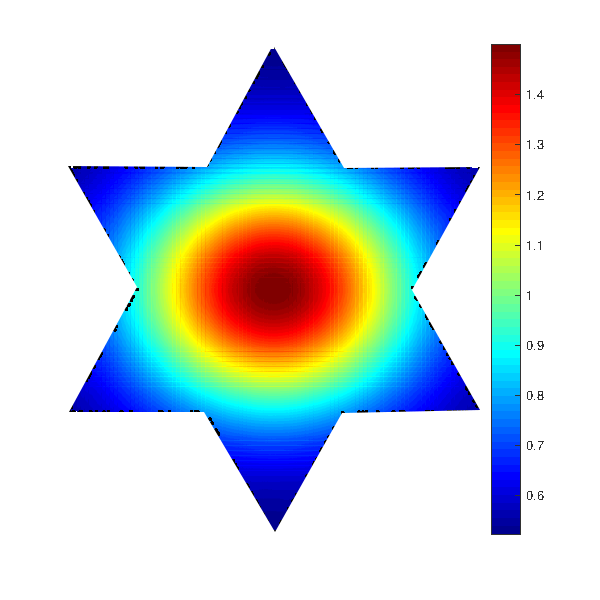}
		
	}\hfill{}\subfloat[\label{fig:PIELM_err_star_diff}Error plot.]{\includegraphics[scale=0.4]{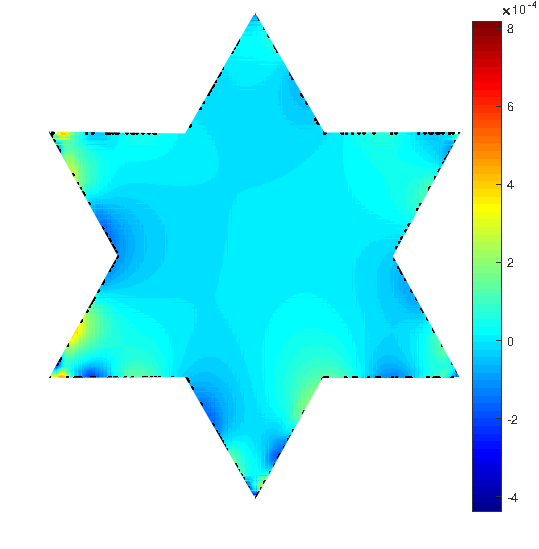}
		
	}
	
	\caption{\label{fig:4.1.6_2D_diffu}Solution and error for 2D steady diffusion
		equation on $\varOmega_{1}$.}
\end{figure}
\begin{figure}[H]
	\subfloat[\label{fig:4.1.8a_sol}PIELM solution. ]{\includegraphics[scale=0.4]{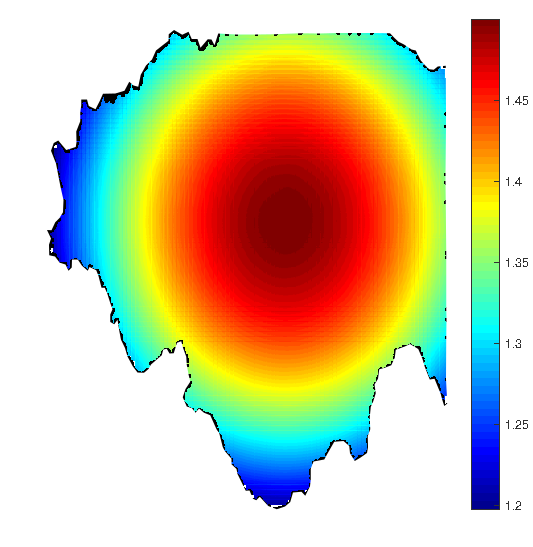}
		
	}\hfill{}\subfloat[\label{fig:4.1.8b_illinois_err}Error plot.]{\begin{raggedleft}
			\includegraphics[scale=0.4]{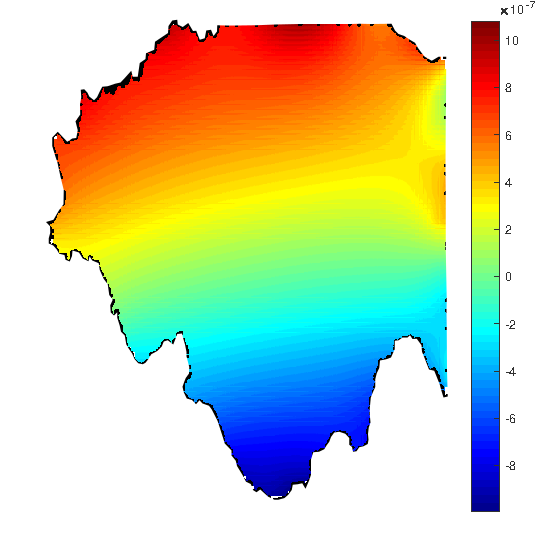}
			\par\end{raggedleft}
	}
	
	\caption{\label{fig:4.1.8_illinois_grid}Solution and error for 2D diffusion
		equation in a complex 2D geometry.}
\end{figure}
The stationary 2D advection and diffusion equations for the three
cases are given by 
\begin{equation}
au_{x}+bu_{y}=R,(x,y)\epsilon\Omega_{1}
\end{equation}

\begin{equation}
u_{xx}+u_{yy}=R,(x,y)\epsilon\Omega_{1}
\end{equation}

\begin{equation}
u_{xx}+u_{yy}=R,(x,y)\epsilon\Omega_{2}
\end{equation}
where $a=1$ and $b=\frac{1}{2}$ are advection coefficients. The
computational domains $\varOmega_{1}$ and $\varOmega_{2}$ are shown
in respectively. The expressions for $R$ and the $B$ for these cases
are constructed by choosing the following exact solutions
\begin{equation}
\widehat{u}=\frac{1}{2}cos(\pi x)sin(\pi y),
\end{equation}

\begin{equation}
\widehat{u}=\frac{1}{2}+e^{-(2x^{2}+4y^{2})},
\end{equation}
\begin{equation}
\widehat{u}=\frac{1}{2}+e^{-((x-0.6)^{2}+(y-0.6)^{2})}
\end{equation}
respectively. PIELM equations for these problems are as follows:
\begin{enumerate}
	\item $\text{\ensuremath{\overrightarrow{\xi}_{f}}=\ensuremath{\overrightarrow{0}}}$
	\begin{itemize}
		\item Case 1: Advection
		\begin{equation}
		\varphi'(\boldsymbol{X_{f}}\boldsymbol{W}^{T})\text{\ensuremath{\odot}}\overrightarrow{c}\text{\ensuremath{\odot}}(a\overrightarrow{m}+b\overrightarrow{n})=R(\overrightarrow{x}_{f},\overrightarrow{y}_{f})
		\end{equation}
		
		\item Case 2: Diffusion
		\begin{equation}
		\varphi''(\boldsymbol{X_{f}}\boldsymbol{W}^{T})\text{\ensuremath{\odot}}\overrightarrow{c}\text{\ensuremath{\odot}}(\overrightarrow{m}\text{\ensuremath{\odot\overrightarrow{m}}}+\overrightarrow{n}\text{\ensuremath{\odot\overrightarrow{n}}})=R(\overrightarrow{x}_{f},\overrightarrow{y}_{f})
		\end{equation}
		where $\boldsymbol{X_{f}}=[\overrightarrow{x}_{f},\overrightarrow{y}_{f},\overrightarrow{I}]$
		and $\boldsymbol{W}=[\overrightarrow{m},\overrightarrow{n},\overrightarrow{b}]$
		(refer Fig (\ref{fig:PiELM_2D_net})).
	\end{itemize}
	\item $\overrightarrow{\xi}_{bc}=\overrightarrow{0}$
	\begin{equation}
	\varphi(\boldsymbol{X_{bc}}\boldsymbol{W}^{T})\overrightarrow{c}=B(\overrightarrow{x}_{bc},\overrightarrow{y}_{bc})
	\end{equation}
	where $\boldsymbol{X_{bc}}=[\overrightarrow{x}_{bc},\overrightarrow{y}_{bc},\overrightarrow{I}]$.
\end{enumerate}
The results for the advection and diffusion cases on $\varOmega_{1}$
are shown in Fig (\ref{fig:8B}) and Fig (\ref{fig:4.1.6_2D_diffu})
respectively. The result for diffusion case on $\varOmega_{2}$ is
given in Fig (\ref{fig:4.1.8_illinois_grid}). Summary of the experiments
is given in 
\begin{table}[H]
	\begin{centering}
		\begin{tabular}{|c|>{\centering}p{3cm}|>{\centering}p{3cm}|}
			\hline 
			TC & $[N_{f},N_{BC},N^{*}]$ & $\mathcal{O}(Error)$\tabularnewline
			\hline 
			\hline 
			TC-4 & $[921,240,2000]$ & $10^{-6}$\tabularnewline
			\hline 
			TC-5 & $[921,240,2000]$ & $10^{-4}$\tabularnewline
			\hline 
			TC-6 & $[1489,881,5000]$ & $10^{-7}$\tabularnewline
			\hline 
		\end{tabular}
		\par\end{centering}
	\caption{\label{tab:2d_steady_results}Summary of experiments for 2D steady
		test cases.}
\end{table}

\subsubsection*{Remarks}
\begin{enumerate}
	\item The unified deep ANN algorithm \cite{BERG ET AL} took $5500$
	points to achieve an order of accuracy of $10^{-3}$ in TC-4 and TC-5.
	In comparison, PIELM solved these cases with merely $1161$ points
	and still achieved an order of accuracy of $10^{-6}$ and $10^{-4}$
	respectively.
	\item False diffusion is an error which gives diffusion like appearance
	to solution of pure advection equation when it is solved using upwind
	discretization. In Fig (\ref{fig:8B}) these errors can be seen flowing
	along streamlines. However, PIELM reduces the order of these errors
	to an insignificant level. 
	\item Computational domain $\Omega_{2}$ is the map of state of Illinois
	in USA. This is an example of a complicated polygon which has very
	short line segments and fine grained details in various regions. Conventional
	mesh based methods are not feasible for these kind of geometries.
	The latitudes and longitudes of the boundary are available in MATLAB's
	in-built function ``usamap''. We have re-scaled the data in the
	range $0-1$. PIELM solved this case with just $2370$ points to an
	accuracy level of $10^{-7}$ in just $10$ seconds. 
\end{enumerate}

\subsection{1D unsteady advection cases {[} TC-7, TC-8 {]}}

\begin{figure}[H]
	\subfloat[\label{fig:adv_const}Constant coefficient unsteady advection.]{\includegraphics[scale=0.7]{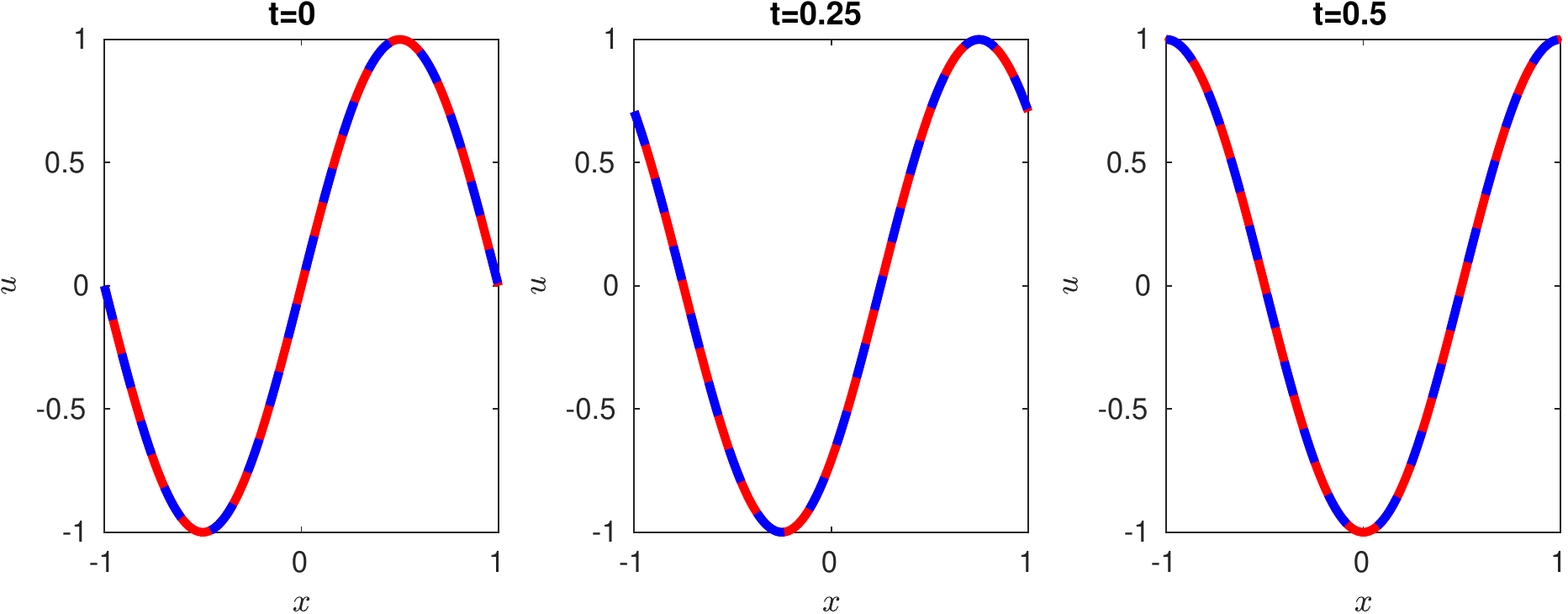}
		
	}\hfill{}\subfloat[\label{fig:adv_var}Variable coefficient unsteady advection.]{\includegraphics[scale=0.7]{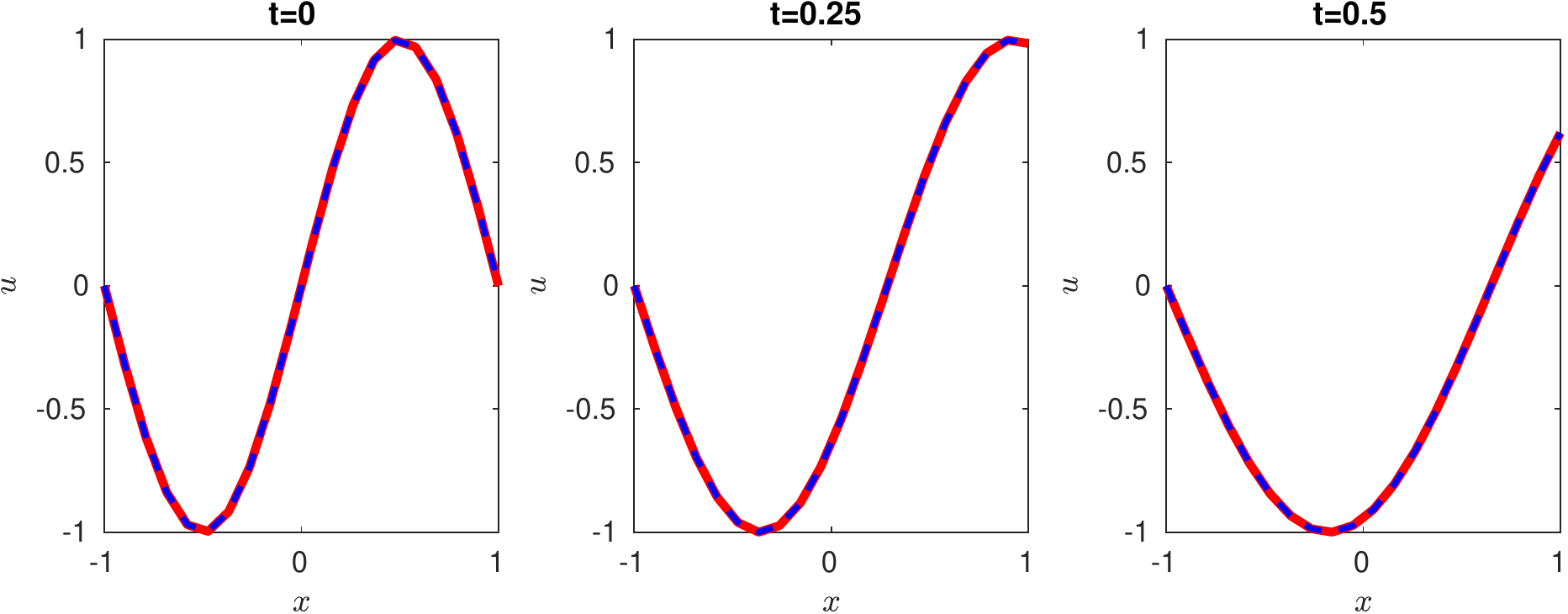}
		
	}
	
	\caption{\label{fig:4.1.2.1}Exact and PIELM solution for 1D unsteady advection
		with constant and variable coefficients. Red: PIELM, Blue: Exact.}
\end{figure}
The unsteady 1D advection equation is given by 
\begin{equation}
u_{t}+a(x)u_{x}=0,(x,t)\epsilon[-1,1]\text{x}[0,0.5],
\end{equation}
\begin{equation}
u(x,0)=F(x),x\epsilon[-1,1],
\end{equation}
where $a(x,t)$ is the advection coefficient. In this problem, we
consider two cases: (1) constant coefficient case with $a(x,t)=1$
and (2) variable coefficient case \cite{KOPRIVA} with $a(x,t)=1+x$.
The two cases have periodic and inflow boundary conditions respectively.
The value of $F$ is $sin(\pi x)$. The exact solutions to the 1D
unsteady advection problems with constant and variable coefficient
are respectively given by

\begin{equation}
\widehat{u}=F(x-t),
\end{equation}

\begin{equation}
\widehat{u}=F((1+x)e^{-t}-1).
\end{equation}

\subsubsection*{PIELM equations }
\begin{enumerate}
	\item $\text{\ensuremath{\overrightarrow{\xi}_{f}}=\ensuremath{\overrightarrow{0}}}$
	\begin{itemize}
		\item TC-7: Linear case
		\begin{equation}
		\varphi'(\boldsymbol{X_{f}}\boldsymbol{W}^{T})\text{\ensuremath{\odot}}\overrightarrow{c}\text{\ensuremath{\odot}}(\overrightarrow{m}+\overrightarrow{n})=\overrightarrow{0}
		\end{equation}
		\item TC-8: Quasi-linear case 
		\begin{equation}
		\varphi'(\boldsymbol{X_{f}}\boldsymbol{W}^{T})\overrightarrow{c}\text{\ensuremath{\odot}}\overrightarrow{n}+(\varphi'(\boldsymbol{X_{f}}\boldsymbol{W}^{T})\text{\ensuremath{\odot}}\boldsymbol{A_{f}})\overrightarrow{c}\text{\ensuremath{\odot}}\overrightarrow{m}=\overrightarrow{0}
		\end{equation}
		where $\overrightarrow{x}_{f},\overrightarrow{t}_{f}$ are collocation
		point vectors, $\boldsymbol{X_{f}}=[\overrightarrow{x}_{f},\overrightarrow{t}_{f},\overrightarrow{I}]$,
		$\boldsymbol{W}=[\overrightarrow{m},\overrightarrow{n},\overrightarrow{b}]$
		and $\boldsymbol{A_{f}}=[\begin{array}{cccc}
		a\left(\overrightarrow{x_{f}},\overrightarrow{t_{f}}\right), & ... & , & a\left(\overrightarrow{x_{f}},\overrightarrow{t_{f}}\right)\end{array}]_{N\text{x}N^{*}}$. 
	\end{itemize}
	\item $\overrightarrow{\xi}_{bc}=\overrightarrow{0}$ 
	\begin{equation}
	\varphi(\boldsymbol{X_{lbc}}\boldsymbol{W}^{T})\overrightarrow{c}=\varphi(\boldsymbol{X_{rbc}}\boldsymbol{W}^{T})\overrightarrow{c}
	\end{equation}
	where $\overrightarrow{x}_{lbc}$, $\overrightarrow{t}_{lbc}$ are
	left boundary points vectors, $\overrightarrow{x}_{rbc}$, $\overrightarrow{t}_{rbc}$
	are right boundary points vectors, $\boldsymbol{X_{lbc}}=[\overrightarrow{x}_{lbc},\overrightarrow{t}_{lbc},\overrightarrow{I}]$
	and $\boldsymbol{X_{rbc}}=[\overrightarrow{x}_{rbc},\overrightarrow{t}_{rbc},\overrightarrow{I}]$.
	\item $\overrightarrow{\xi}_{ic}=\overrightarrow{0}$ 
	\begin{equation}
	\varphi(\boldsymbol{X_{ic}}\boldsymbol{W}^{T})\overrightarrow{c}=F(\overrightarrow{x}_{ic},\overrightarrow{t}_{ic})
	\end{equation}
	where$\overrightarrow{x}_{ic},\overrightarrow{t}_{ic}$ are initial
	condition vectors, $\boldsymbol{X_{ic}}=[\overrightarrow{x}_{ic},\overrightarrow{t}_{ic},\overrightarrow{I}]$
	and $F$ is initial condition.
\end{enumerate}
The results are shown in Fig (\ref{fig:4.1.2.1}). PIELM predicts
the exact solution correctly for both linear and quasi-linear advection.
In this case, we took $N_{f}=420$, $N_{bc}=21$, $N_{ic}=20$ and
$N^{*}=440$. The total learning time is within 2-3 seconds.

\subsubsection*{Remark}
\begin{enumerate}
	\item For advection problems, the time step of the traditional numerical
	schemes like upwinding can not exceed the mesh size due to stability
	issues. However, PIELM doesn't impose any such restriction and we
	can take larger time steps.
\end{enumerate}

\subsubsection{1D unsteady advection-diffusion {[} TC-9 {]}}

\begin{figure}[H]
	\begin{centering}
		\includegraphics[scale=0.65]{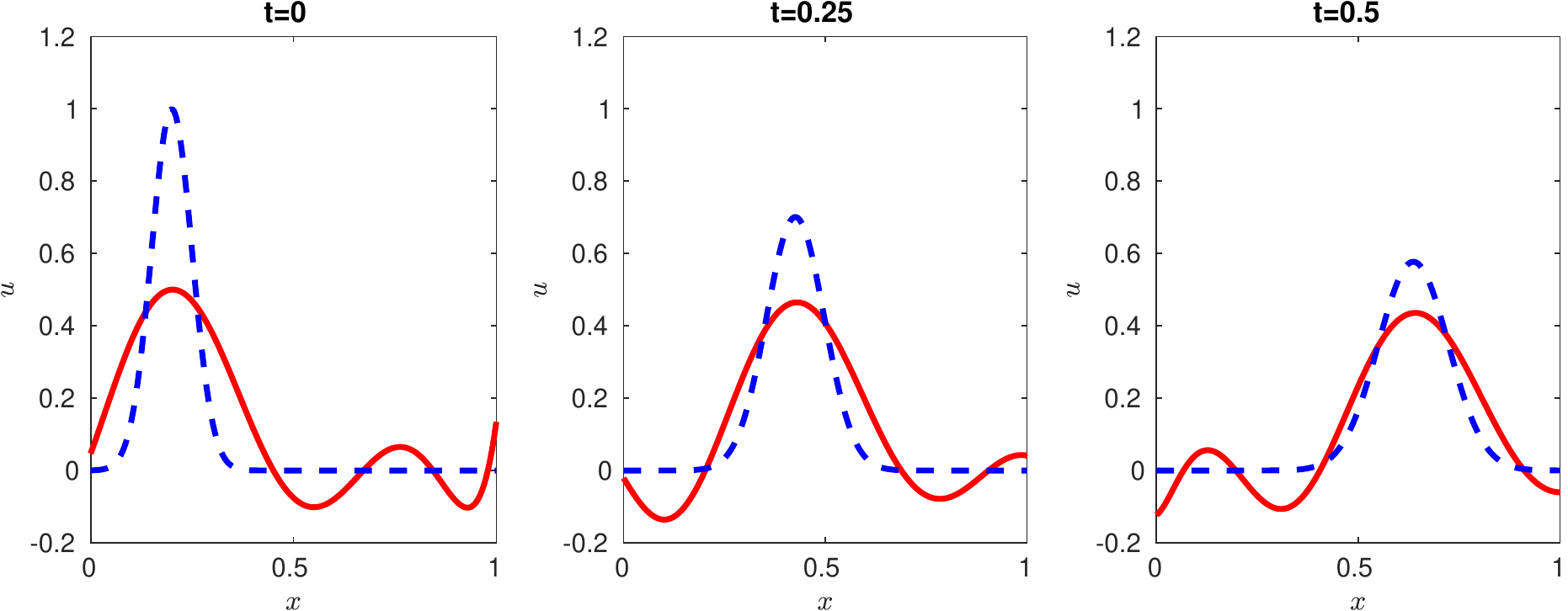}
		\par\end{centering}
	\caption{\label{fig:4.1.2.2}Exact and PIELM solution for unsteady 1D convection
		diffusion at $\nu=0.005$. Red: PIELM, Blue: Exact.}
\end{figure}
The 1D equivalent of the unsteady 2D advection-diffusion equation
solved by Borker et al. \cite{FARHAT} is given by 
\begin{equation}
u_{t}+au_{x}=\nu u_{xx},(x,t)\;\epsilon\;[0,1]\text{x}[0,0.5],
\end{equation}
where $a$ is the advection coefficient. The expressions for the initial
condition $F$ and the boundary condition $B$ are constructed on
the basis of the following exact solution
\begin{equation}
\widehat{u}=\frac{1}{\sqrt{4t+1}}e^{-\frac{1}{\nu(4t+1)}(x-0.2-at){}^{2}}
\end{equation}
The PIELM equations are as follows: 
\begin{enumerate}
	\item $\text{\ensuremath{\overrightarrow{\xi}_{f}}=\ensuremath{\overrightarrow{0}}}$
	\begin{equation}
	\varphi'(\boldsymbol{X_{f}}\boldsymbol{W}^{T})\overrightarrow{c}\text{\ensuremath{\odot}}\overrightarrow{n}+a\varphi'(\boldsymbol{X_{f}}\boldsymbol{W}^{T})\overrightarrow{c}\text{\ensuremath{\odot}}\overrightarrow{m}=\nu\varphi''(\boldsymbol{X_{f}}\boldsymbol{W}^{T})\text{\ensuremath{\odot}}\overrightarrow{c}\text{\ensuremath{\odot}}\overrightarrow{m}\text{\ensuremath{\odot\overrightarrow{m}}}
	\end{equation}
	where $\overrightarrow{x}_{f},\overrightarrow{t}_{f}$ are collocation
	point vectors, $\boldsymbol{X_{f}}=[\overrightarrow{x}_{f},\overrightarrow{t}_{f},\overrightarrow{I}]$
	and $\boldsymbol{W}=[\overrightarrow{m},\overrightarrow{n},\overrightarrow{b}]$.
	\item $\overrightarrow{\xi}_{bc}=\overrightarrow{0}$ 
	\begin{equation}
	\varphi(\boldsymbol{X_{bc}}\boldsymbol{W}^{T})\overrightarrow{c}=B(\overrightarrow{x}_{bc},\overrightarrow{t}_{bc})
	\end{equation}
	where $\overrightarrow{x}_{bc}$, $\overrightarrow{t}_{bc}$ are boundary
	points vectors, $\boldsymbol{X_{bc}}=[\overrightarrow{x}_{bc},\overrightarrow{t}_{bc},\overrightarrow{I}]$
	and $B$ is the boundary condition. 
	\item $\overrightarrow{\xi}_{ic}=\overrightarrow{0}$ 
	\begin{equation}
	\varphi(\boldsymbol{X_{ic}}\boldsymbol{W}^{T})\overrightarrow{c}=F(\overrightarrow{x}_{ic},\overrightarrow{t}_{ic})
	\end{equation}
	where$\overrightarrow{x}_{ic},\overrightarrow{t}_{ic}$ are initial
	condition vectors, $\boldsymbol{X_{ic}}=[\overrightarrow{x}_{ic},\overrightarrow{t}_{ic},\overrightarrow{I}]$
	and $F$ is initial condition.
\end{enumerate}
The results are shown in Fig (\ref{fig:4.1.2.2}). In this case, PIELM
prediction clearly goes wrong in the following grounds:
\begin{enumerate}
	\item Initial and boundary conditions aren't captured correctly.
	\item Solution has unphysical oscillations throughout the domain. 
	\item The hump of the Gaussian decays a lot slower than the correct rate.
\end{enumerate}
We increased the size of hidden layer but didn't see any improvement
in results. For example, we took $250\text{x}30$ points in the computational
domain and put as many as $7780$ neurons in the hidden layer. Still
the PIELM predictions were poor. 

\subsubsection{2D unsteady advection-diffusion {[} TC-10 {]}}

\begin{figure}[H]
	\begin{centering}
		\includegraphics[scale=0.1]{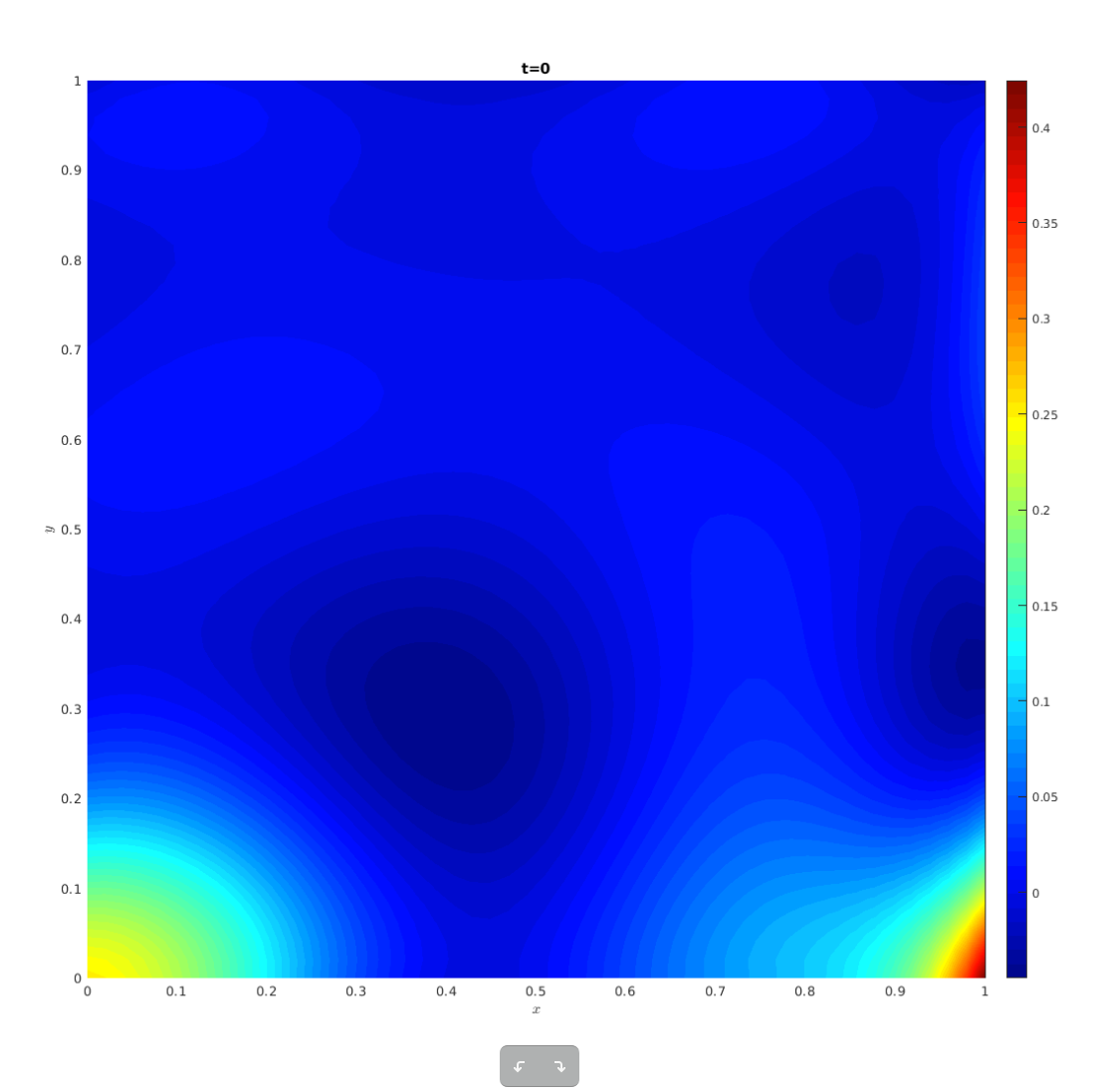}\includegraphics[scale=0.1]{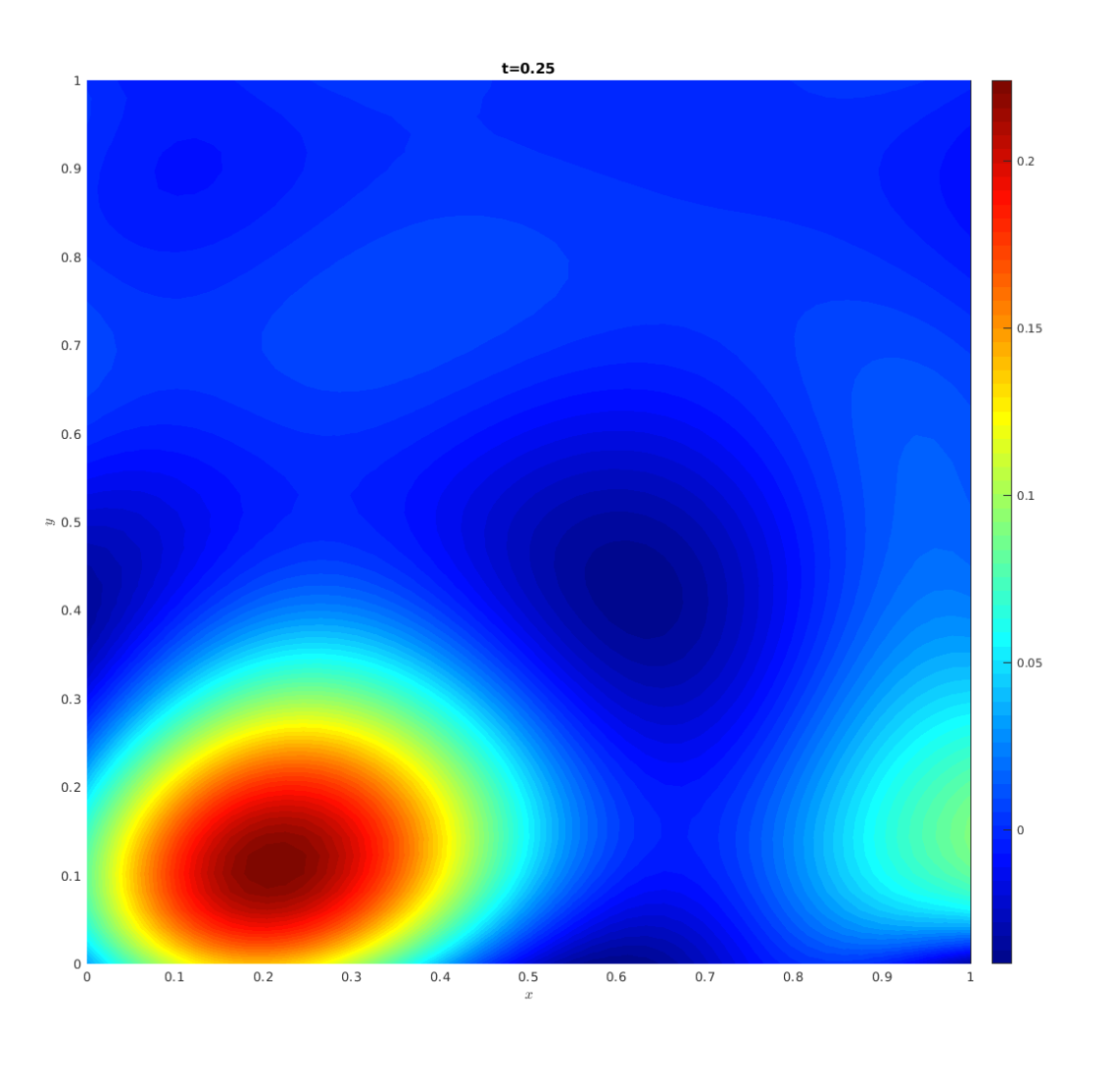}\includegraphics[scale=0.1]{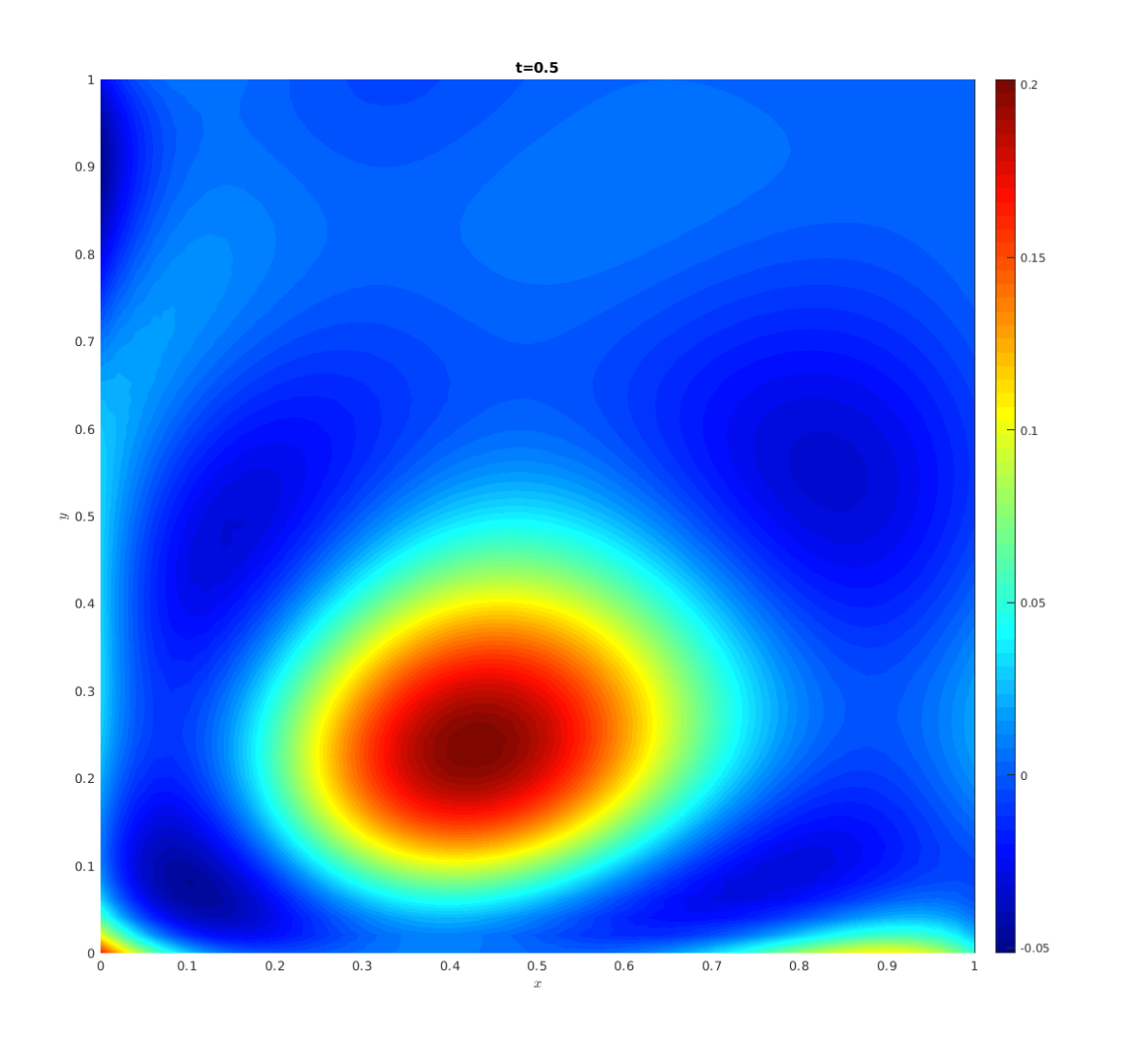}
		\par\end{centering}
	\caption{\label{fig:pielm_2d_cd_sol}PIELM solution for 2D unsteady advection
		diffusion}
\end{figure}
\begin{figure}[H]
	\begin{centering}
		\includegraphics[scale=0.1]{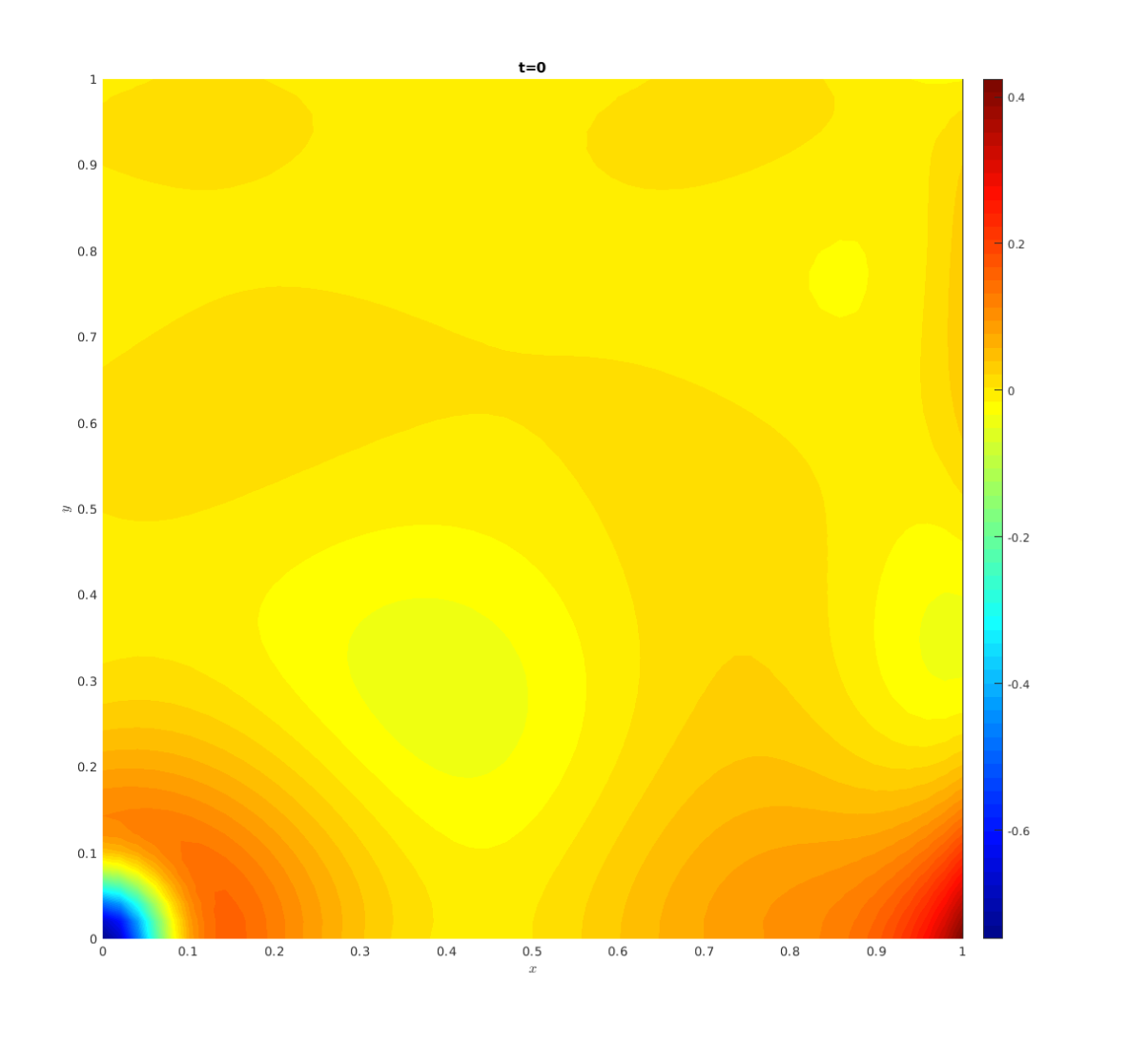}\includegraphics[scale=0.1]{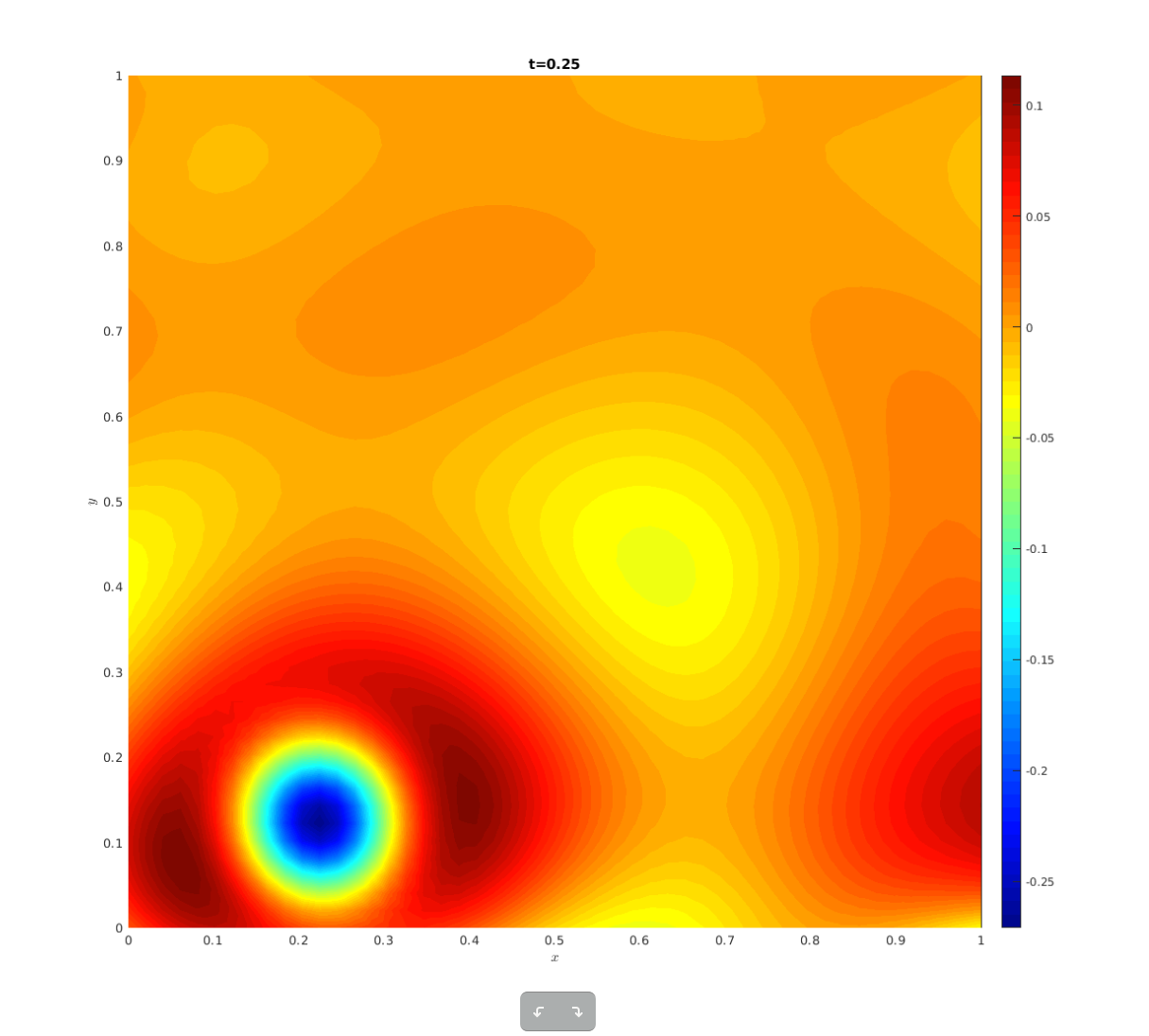}\includegraphics[scale=0.1]{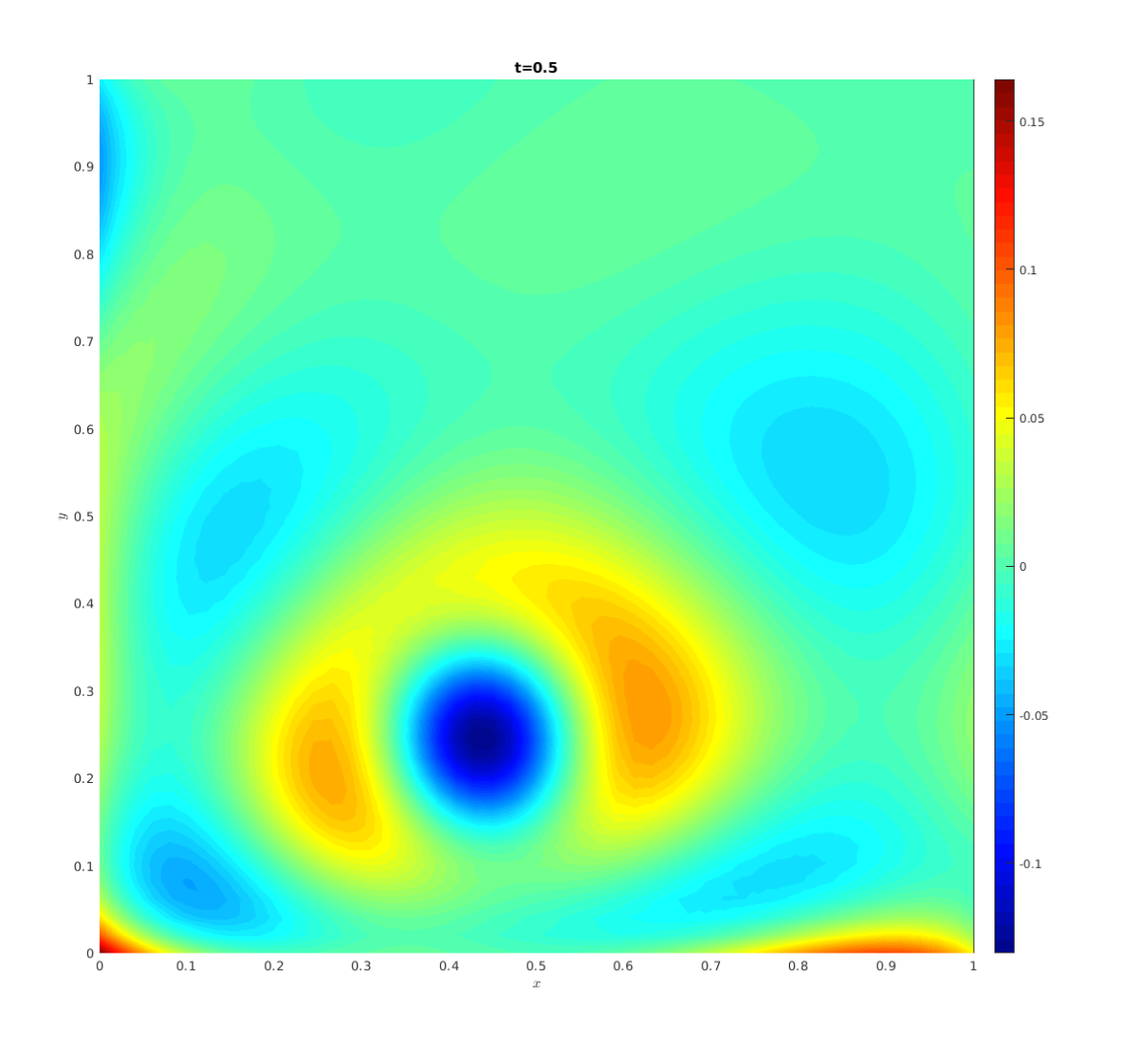}
		\par\end{centering}
	\caption{\label{fig:pielm_2d_cd_err}Error for 2D unsteady advection diffusion}
\end{figure}
We solve the following unsteady 2D advection-diffusion equation\cite{FARHAT}
\begin{equation}
u_{t}+au_{x}+bu_{y}=\nu(u_{xx}+u_{yy}),(x,y,t)\epsilon\Omega_{xy}\text{x}[0,0.2],
\end{equation}
\begin{equation}
u(x,y,0)=F(x,y),(x,y)\epsilon\Omega_{xy},
\end{equation}
\begin{equation}
u(x,y,t)=B(x,y,t),(x,y,t)\epsilon\partial\Omega_{xy}\text{x}[0,0.5],
\end{equation}
where $\Omega_{xy}=[0,1]\text{x}[0,1]$. $a=cos(22.5)$ and $b=sin(22.5)$
are advection coefficients in $x$ and $y$ directions and $\nu=0.005$
is the diffusion coefficient. The expressions for the initial and
the boundary conditions are constructed by choosing the following
exact solution:
\begin{equation}
\widehat{u}=\frac{1}{(4t+1)}e^{-\frac{(x-at)^{2}+(y-bt)^{2}}{\nu(4t+1)}}
\end{equation}
PIELM equations to be solved are as follows:
\begin{enumerate}
	\item $\text{\ensuremath{\overrightarrow{\xi}_{f}}=\ensuremath{\overrightarrow{0}}}$
	\begin{equation}
	\varphi'(\boldsymbol{X_{f}}\boldsymbol{W}^{T})\overrightarrow{c}\text{\ensuremath{\odot}}\overrightarrow{r}+\varphi'(\boldsymbol{X_{f}}\boldsymbol{W}^{T})\overrightarrow{c}\text{\ensuremath{\odot}}(a\overrightarrow{p}+b\overrightarrow{q})=\nu\varphi''(\boldsymbol{X_{f}}\boldsymbol{W}^{T})\text{\ensuremath{\odot}}\overrightarrow{c}\text{\ensuremath{\odot}}(\overrightarrow{p}\text{\ensuremath{\odot\overrightarrow{p}}+\ensuremath{\overrightarrow{q}\text{\ensuremath{\odot\overrightarrow{q}}}})}
	\end{equation}
	where $\overrightarrow{x}_{f},\overrightarrow{y}_{f},\overrightarrow{t}_{f}$
	are collocation point vectors, $\boldsymbol{X_{f}}=[\overrightarrow{x}_{f},\overrightarrow{y}_{f},\overrightarrow{t}_{f},\overrightarrow{I}]$.
	Referring to Fig (\ref{fig:PIELM-2D-unsteady}), $\boldsymbol{W}=[\overrightarrow{p},\overrightarrow{q},\overrightarrow{r},\overrightarrow{s}]$.
	\item $\overrightarrow{\xi}_{bc}=\overrightarrow{0}$ 
	\begin{equation}
	\varphi(\boldsymbol{X_{bc}}\boldsymbol{W}^{T})\overrightarrow{c}=B(\overrightarrow{x}_{bc},\overrightarrow{y}_{bc},\overrightarrow{t}_{bc})
	\end{equation}
	where $\overrightarrow{x}_{bc}$, $\overrightarrow{y}_{bc}$, $\overrightarrow{t}_{bc}$
	are boundary points vectors and $\boldsymbol{X_{bc}}=[\overrightarrow{x}_{bc},\overrightarrow{y}_{bc},\overrightarrow{t}_{bc},\overrightarrow{I}]$.
	\item $\overrightarrow{\xi}_{ic}=\overrightarrow{0}$ 
	\begin{equation}
	\varphi(\boldsymbol{X_{ic}}\boldsymbol{W}^{T})\overrightarrow{c}=F(\overrightarrow{x}_{ic},\overrightarrow{y}_{ic},\overrightarrow{t}_{ic})
	\end{equation}
	where$\overrightarrow{x}_{ic},\overrightarrow{y}_{ic},\overrightarrow{t}_{ic}$
	are initial condition vectors and $\boldsymbol{X_{ic}}=[\overrightarrow{x}_{ic},\overrightarrow{y}_{ic},\overrightarrow{t}_{ic},\overrightarrow{I}]$.
\end{enumerate}
The results are shown in Fig (\ref{fig:pielm_2d_cd_sol}-\ref{fig:pielm_2d_cd_err}). They are 2D equivalents of 1D results. The solution is diffusive and contains unphysical oscillations throughout the domain. In spite of all these issues, it should be noted that: 
\begin{enumerate}
	\item The advection component of the equation is captured correctly. The
	two humps are moving at the same speed. 
	\item The non-physical oscillations don't grow with time.
\end{enumerate}
For this case, we have taken a total of 125000 data points and 1000 neurons in the hidden layer. The results show signs of improvement if we keep increasing the number of points.
However, that takes a lot of time which makes the option impractical.
The limitations of the PIELM will be further investigated in the next
section. We close this section by summarizing the advantages of the
PIELM which are as follows:
\begin{enumerate}
	\item It is extremely fast as well as data efficient.(TC-1 to TC-6) 
	\item It can be seamlessly extended to higher dimensions. (TC-10)
	\item It reduces numerical artefacts like false diffusion.(TC-4)
	\item It is meshfree method and can handle the complex geometries.(TC-6)
	\item It reduces the arbitrariness of the number of neurons in the hidden
	layer. 
\end{enumerate}
\section{\label{sec:5}Limitations of PIELM}

\begin{table}[H]
	\begin{tabular}{|>{\raggedright}p{4cm}|>{\centering}p{2cm}|>{\centering}p{5cm}|}
		\hline 
		& Test case ID & Description\tabularnewline
		\hline 
		\hline 
		\multirow{2}{4cm}{Representation of functions with PIELM} & TC-11 & Representation of sharp and discontinuous functions in 1D\tabularnewline
		\cline{2-3} 
		& TC-12 & Representation of a sharp peaked 2D Gaussian \tabularnewline
		\hline 
		\multirow{2}{4cm}{Solution of PDEs with PIELM} & TC-13 & TC-7 with $F=e^{-100x^{2}}$\tabularnewline
		\cline{2-3} 
		& TC-14 & TC-3 with $\nu=0.02$\tabularnewline
		\hline 
		Solution of PDEs with PINN & TC-15 & TC-7 with $F=e^{-5x^{2}}sin(10\pi x)$\tabularnewline
		\hline 
	\end{tabular}
	
	\caption{\label{tab:limit_PIELM}List of numerical experiments that show limitations
		of PIELM and PINN}
	
\end{table}
A potential reason for the failure of PIELM in solving advection-diffusion
equation \cite{FARHAT} could be the limited representation capacity
of PIELM to represent a complex function. A PDE consists of function
and its derivatives. If a neural network cannot represent the function
itself, then calculation of derivatives only adds to the error. 

\subsubsection*{Representation of functions with sharp gradients {[} TC-11, TC-12
	{]}}

\begin{figure}[H]
	\begin{centering}
		\includegraphics[scale=0.4]{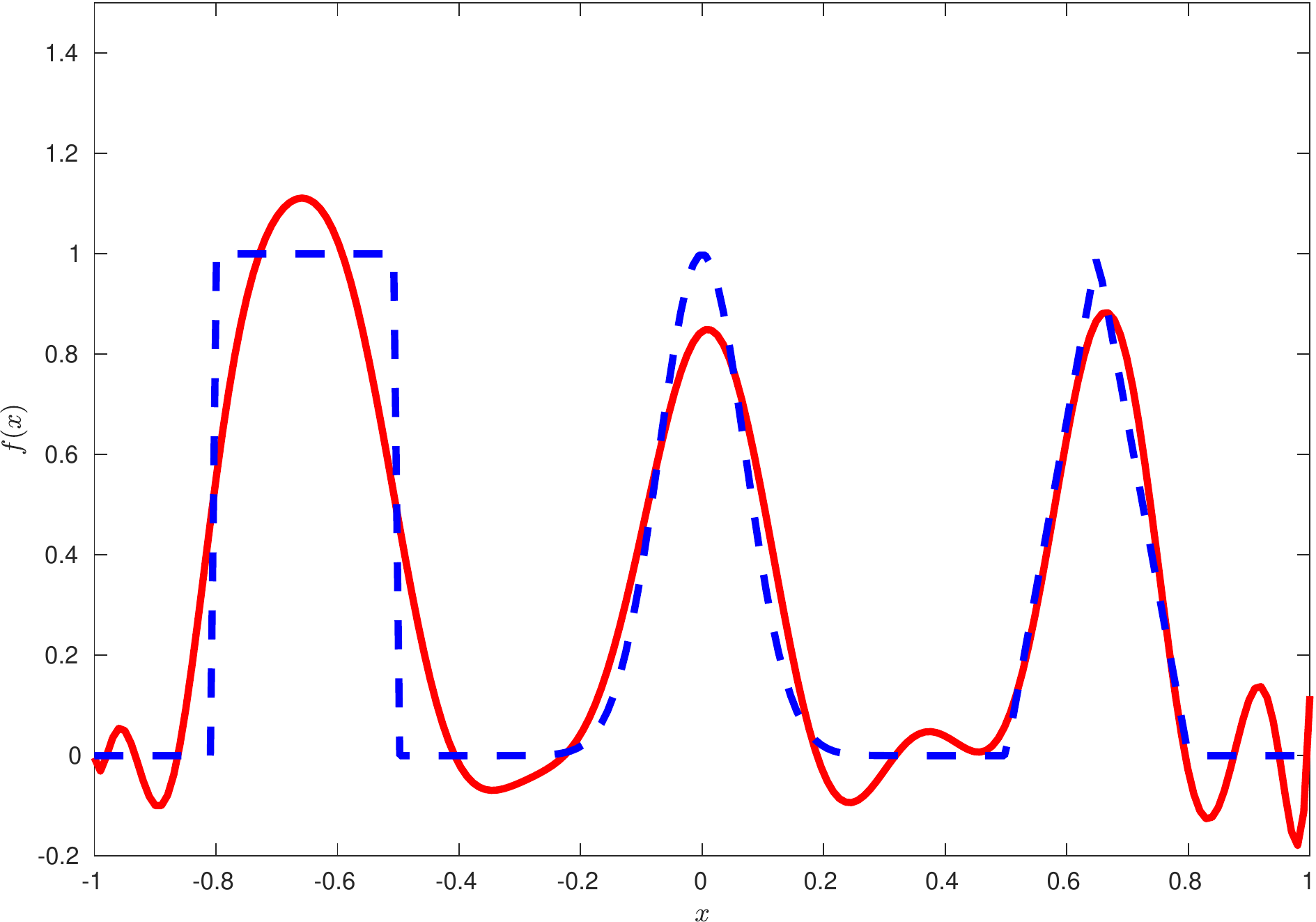}
		\par\end{centering}
	\caption{\label{fig:1D_rep}Representation of 1D non smooth functions with
		PIELM. Red: PIELM solution, Blue: exact solution.}
\end{figure}
\begin{figure}[H]
	\begin{raggedright}
		\includegraphics[scale=0.45]{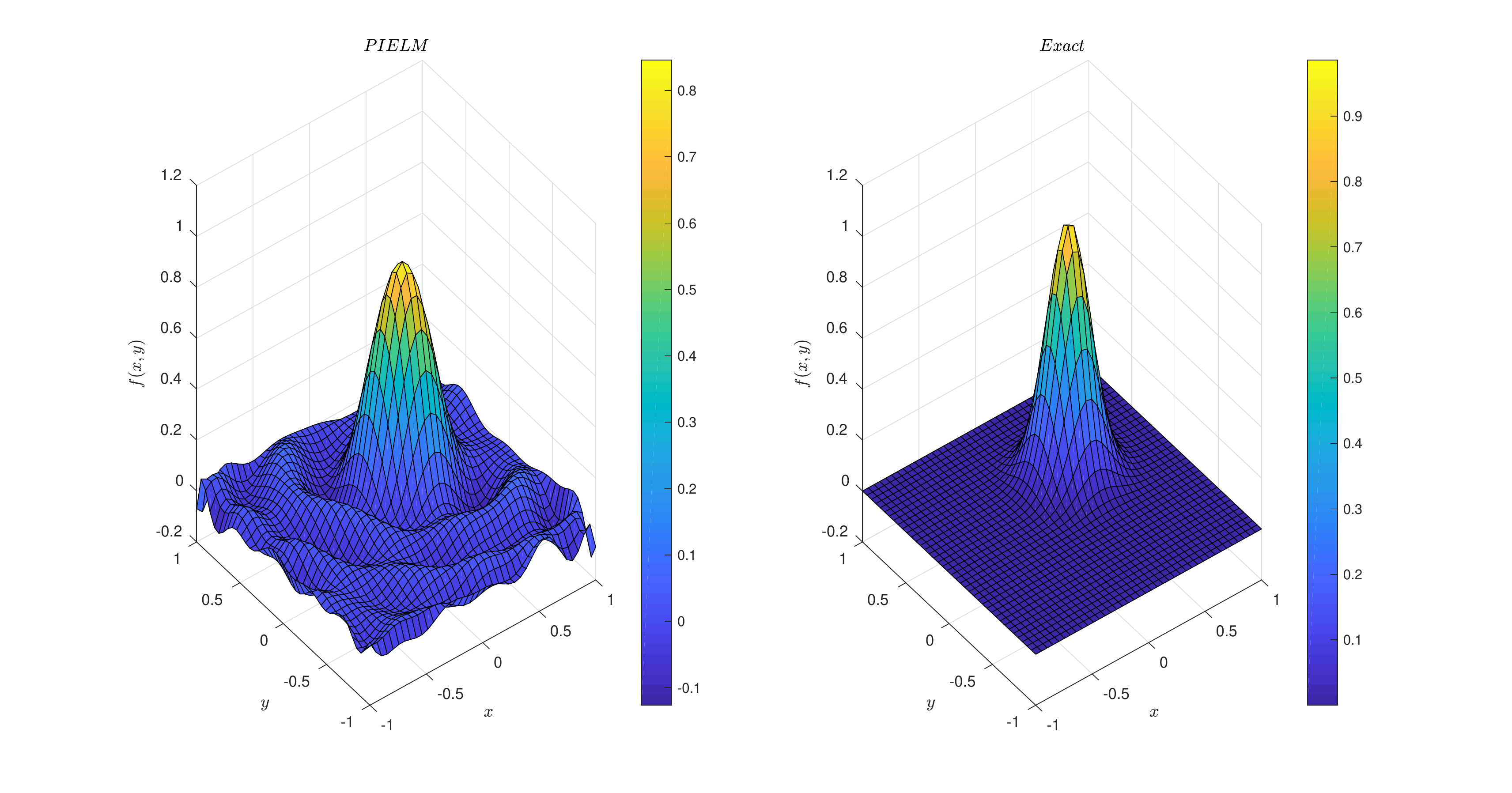}
		\par\end{raggedright}
	\caption{\label{fig:3D_rep} Representation of a sharp peaked 2D Gaussian with
		PIELM }
\end{figure}
To put the representation power of PIELM to test we choose the following
two functions: (1) A 1D composite function that contains both discontinuous
functions and functions with sharp gradients and (2) A 2D Gaussian
function with a sharp peak. The expressions for 2D Gaussian function
and 1D composite function are $f(x,y,t)=e^{-20\{(x-0.25)^{2}+(y-0.25)^{2}\}}$
and 
\begin{equation}
f(x)=\begin{cases}
\frac{1}{2}\{sgn(x+0.8)-sgn(x+0.5)\} & x\leq-\frac{1}{2}\\
e^{-100x^{2}} & -\frac{1}{2}<x\leq\frac{1}{2}\\
\frac{20}{3}x-\frac{10}{3} & \frac{1}{2}<x\leq\frac{13}{20}\\
-\frac{20}{3}x+\frac{16}{3} & \frac{13}{20}<x\leq\frac{4}{5}\\
0 & x>\frac{4}{5}
\end{cases}
\end{equation}
respectively. The results are shown in figure (\ref{fig:1D_rep})and
Fig (\ref{fig:3D_rep}) respectively. These cases clearly expose the
limitation of PIELM in representing profiles with sharp gradients
and corners.

\subsubsection*{PDEs with sharp solutions {[} TC-13, TC-14 {]}}

\begin{figure}[H]
	\begin{centering}
		\includegraphics[scale=0.8]{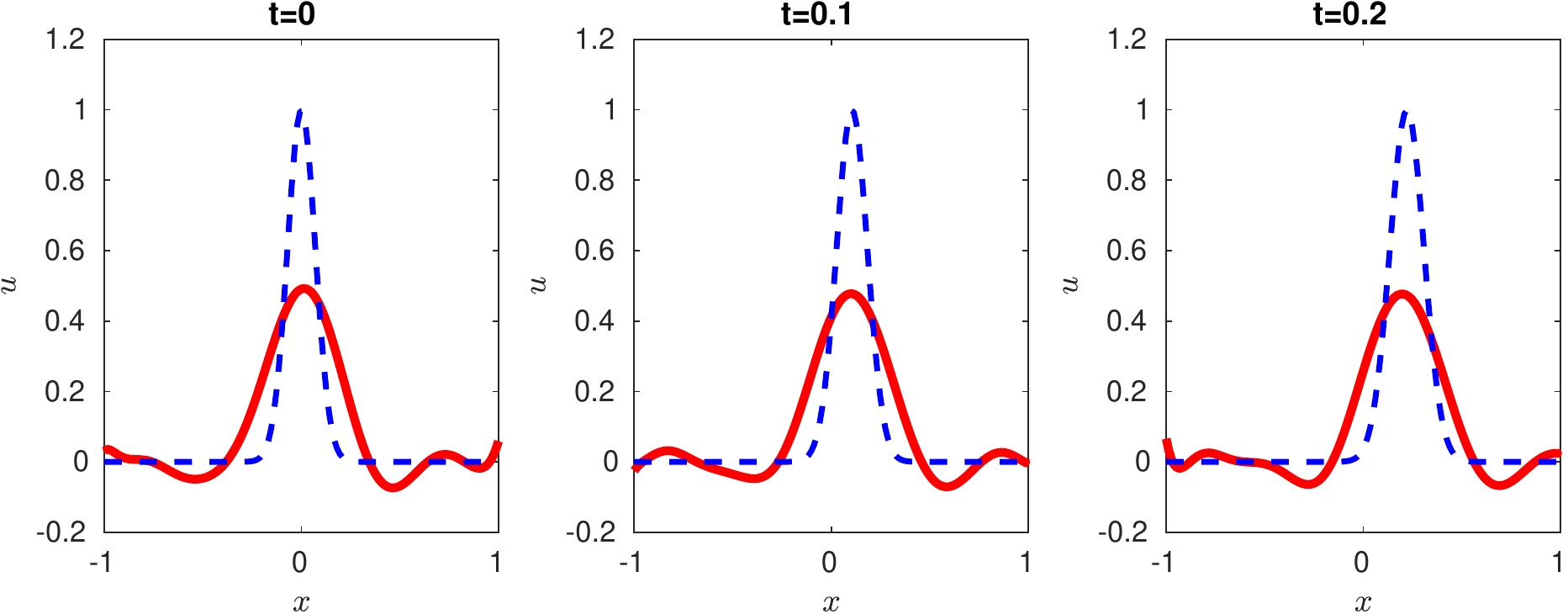}
		\par\end{centering}
	\caption{\label{fig:PIELM_1D_bell_adv}Exact and PIELM solution of 1D advection
		of a sharp peaked Gaussian with PIELM. Red: PIELM, Blue: exact.}
\end{figure}
\begin{figure}[H]
	\begin{centering}
		\includegraphics[scale=0.65]{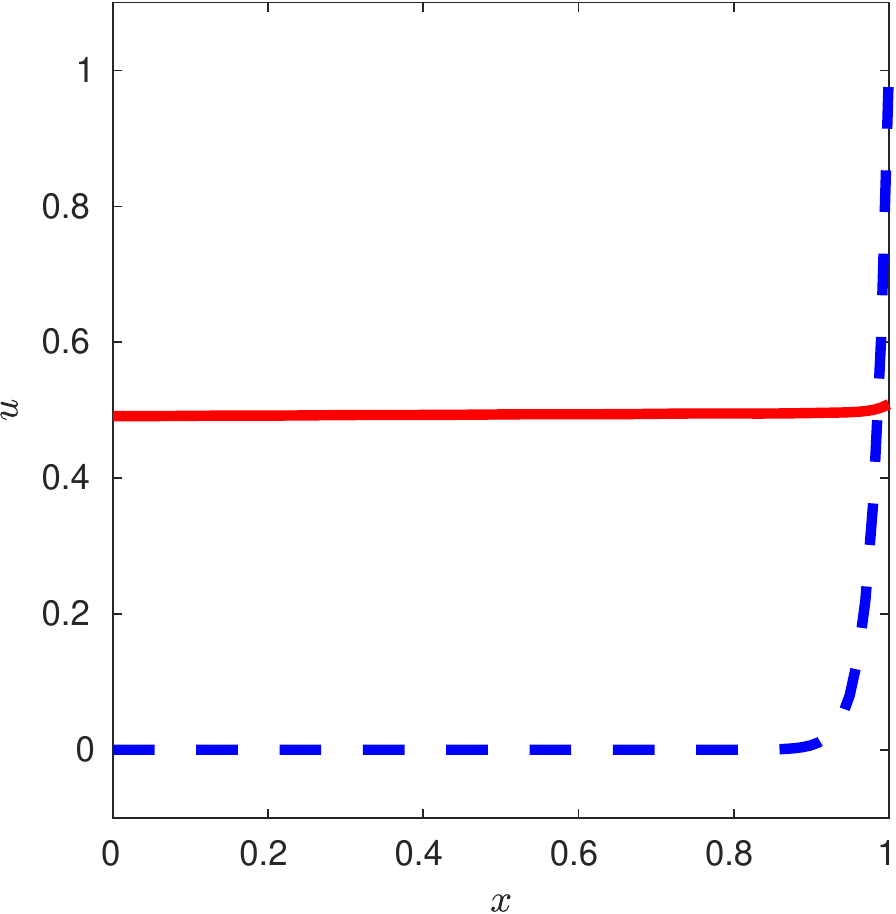}
		\par\end{centering}
	\caption{\label{fig:1D_steady_convection_diffusion_0.02}Steady 1D convection
		diffusion at $\nu=0.02$. Red: PIELM, Blue: Exact.}
\end{figure}
Due to this limitation, PIELM fails to solve any PDE which admits
functions with sharp gradients. For example, we have already seen
the failure of our algorithm in solving 1D and 2D advection-diffusion
equation. We further illustrate the impact of this limitation on two
simpler equations. Firstly, we consider pure advection of a sharp
peaked Gaussian. This equation has been solved in TC-7 for a smooth
sine function. Next , we take steady advection-diffusion equation
(TC-3) with a low value of diffusion coefficient.

The exact and PIELM solutions for these problems are shown in Fig
(\ref{fig:PIELM_1D_bell_adv}) and Fig (\ref{fig:1D_steady_convection_diffusion_0.02})
respectively. 

\subsubsection*{Representation of a high frequency wavelet with PINN {[}TC-15{]}}

\begin{figure}[H]
	\begin{centering}
		\includegraphics[scale=0.3]{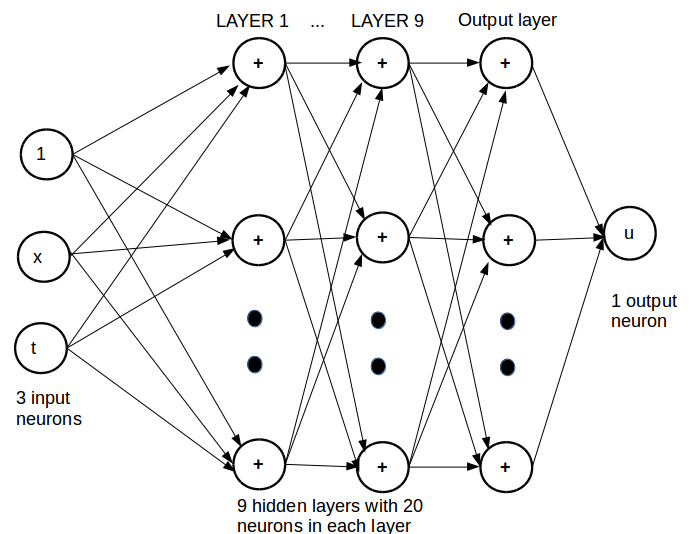}
		\par\end{centering}
	\caption{\label{fig:Raisi_net}Deep PINN architecture}
\end{figure}

\begin{figure}[H]
	\begin{centering}
		\includegraphics[scale=0.5]{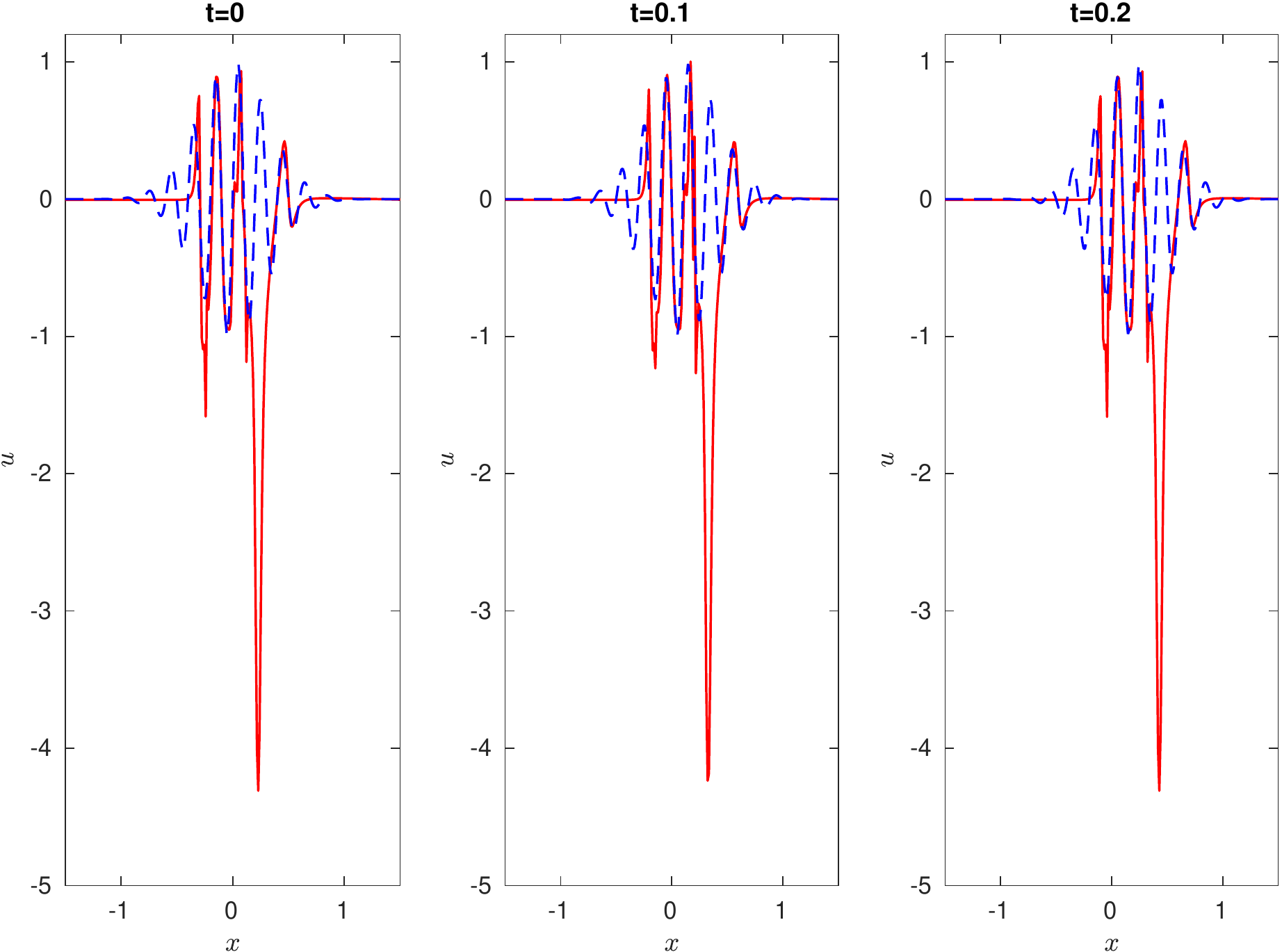}
		\par\end{centering}
	\caption{\label{fig:wave_packet}Exact and PINN solution of pure advection
		of a high frequency wave packet. Red: PINN, Blue: exact}
\end{figure}
An obvious idea to improve the representation capacity of a neural
network is to make it deep. Therefore, we test the performance of
PINN in TC-7 with $F=e^{-5x^{2}}sin(10\pi x)$ i.e. pure advection
of a high frequency wavelet. The PINN code is freely available at
\href{https://github.com/maziarraissi/PINNs.}{https://github.com/maziarraissi/PINNs.}
We modified the original code by replacing the Burgers equation with
pure linear advection equation. Fig(\ref{fig:Raisi_net}) shows the
architecture of PINN. It consists of 9 hidden layers with 20 neurons
each. Each layer is activated by $\tanh$ functions.

The exact and the PINN solutions are shown in Fig (\ref{fig:wave_packet}).
We can see that even a deep network is unable to capture the sharp
gradients of this wavelet. 

List of the experiments conducted in this section are given in Tab
\ref{tab:limit_PIELM}. We summarize this section as follows:
\begin{enumerate}
	\item The main limitation of a PIELM is its inability to represent complex
	functions. 
	\item This limitation restricts the PIELM to solve the PDEs with sharp exact
	solutions.
	\item Adding extra layers is not the practical solution to this problem.
	(TC-15)
\end{enumerate}

\section{\label{sec:6}Distributed PIELM}

In this section, we propose a distributed version of PIELM called
DPIELM. This algorithm takes motivation from finite volume methods
in which the whole computational domain is partitioned into multiple
cells and governing equations are solved at each cell. The solutions
of these individual cells are stitched together with additional convective
and diffusive fluxes conditions at the cell interfaces. We adopt a
similar strategy in DPIELM. As representation of a complex function
is very hard for a single PIELM or PINN in the whole domain, we divide
the domain into multiple cells and install a PIELM in each cell. Therefore,
each PIELM uses different representations in different portions of
domain while satisfying some additional constraints of continuity
and differentiability. 

\subsection{Mathematical formulation}

\begin{figure}[H]
	\subfloat[Distributed PIELMs in $\Omega$]{\begin{raggedright}
			\includegraphics[scale=0.3]{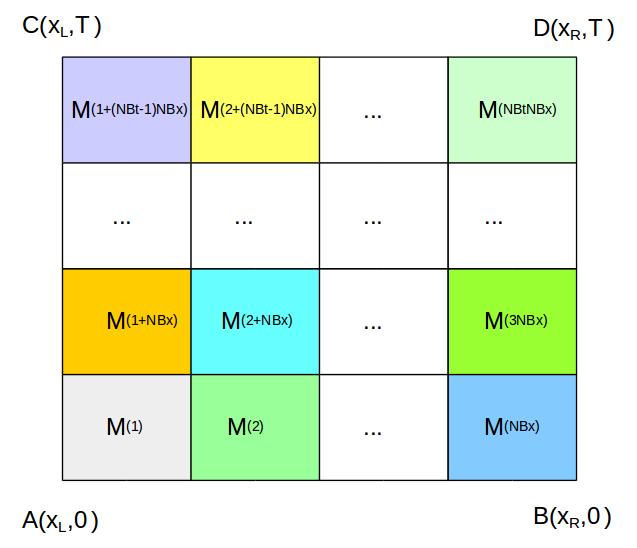}
			\par\end{raggedright}
	}\hfill{}\subfloat[A PIELM in $\Omega_{i}$. Red triangles: boundary points, green rectangles:
	collocation points ]{\begin{raggedleft}
			\includegraphics[scale=0.3]{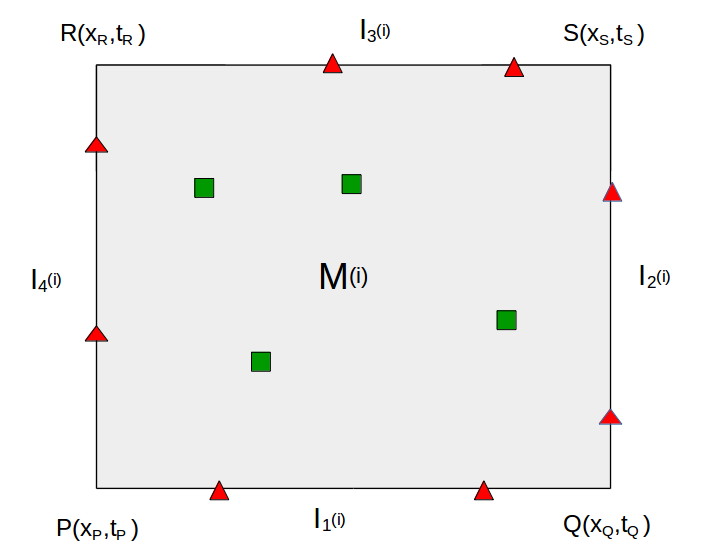}
			\par\end{raggedleft}
	}
	
	\caption{\label{fig:DPIELM in 2D}DPIELM architecture for full domain and an
		individual cell. }
\end{figure}
Consider the following 1D unsteady problem
\begin{equation}
\frac{\partial}{\partial t}u(x,t)+\mathcal{L}u(x,t)=R(x,t),(x,t)\epsilon\varOmega
\end{equation}
\begin{equation}
u(x,t)=B(x,t),(x,t)\epsilon\partial\varOmega,
\end{equation}

\begin{equation}
u(x,0)=F(x),x\epsilon[x_{L},x_{R}],
\end{equation}
where $\mathcal{L}$ is a linear differential operator and $\partial\varOmega$
is the boundary of computational domain $\Omega$. In this problem,
the rectangular domain $\Omega$ is given by $\Omega=[x_{L},x_{R}]\text{x}[0,T]$.
On uniformly dividing $\Omega$ into $N_{c}$ non-overlapping rectangular
cells, $\Omega$ may be written as
\begin{equation}
\Omega=\bigcup_{i=1}^{N_{c}}\Omega_{i}.
\end{equation}
The boundary of the cell $\Omega_{i}$ is denoted by $\partial\varOmega_{i}$.
For rectangular cells,
\begin{equation}
\partial\Omega_{i}=\bigcup_{m=1}^{4}I_{m}^{(i)}
\end{equation}
where $I_{m}^{(i)}$ represents the $m^{th}$ interface of $\Omega_{i}$. 

Fig (\ref{fig:DPIELM in 2D}a) shows the distribution of PIELMs in
a rectangular computational domain with $NB_{x}\text{x}NB_{t}$ cells
(i.e. $N_{c}=NB_{x}\text{x}NB_{t}$). We denote the PIELM on the $i^{th}$
cell by $M^{(i)}$. Fig (\ref{fig:DPIELM in 2D}b) shows a PIELM with
collocation points at the interior and the boundary points at the
four interfaces. The weights and output corresponding to a given $M^{(i)}$
are denoted by $\text{[\ensuremath{\overrightarrow{m}}}^{(i)},\text{\ensuremath{\overrightarrow{n}}}^{(i)},\text{\ensuremath{\overrightarrow{b}}}^{(i)},\text{\ensuremath{\overrightarrow{c}}}^{(i)}]$
and $f^{(i)}$respectively. 

At each $M^{(i)}$, we enforce additional constraints of continuity
(or smoothness) of solution at the cell interfaces depending on the
differential operator $\mathcal{L}$. For example, continuity of solution
is sufficient for advection problems. For diffusion problem, the solution
should be continuously differentiable. For the computational domain
shown in the Fig (\ref{fig:DPIELM in 2D}), the system of equations
to be solved in the DPIELM framework are given below. 

\subsubsection{Regular PIELM equations}
\begin{enumerate}
	\item $\text{\ensuremath{\overrightarrow{\xi}_{f}^{(i)}}=\ensuremath{\overrightarrow{0}} }$
	\begin{equation}
	\left(\frac{\partial\overrightarrow{f}^{(i)}}{\partial t}+\mathcal{L}\overrightarrow{f}^{(i)}-\overrightarrow{R}^{(i)}\right)_{\Omega_{i}}=\overrightarrow{0},i=1,2,...,N_{c}\label{eq:DPIELM_start}
	\end{equation}
	
	\item $\overrightarrow{\xi}_{bc}^{(i)}=\overrightarrow{0}$ 
	\begin{equation}
	(\overrightarrow{f}^{(i)}-\overrightarrow{B}^{(i)})_{I_{4}}=\overrightarrow{0},i=1,(1+NB_{x}),...,(1+(NB_{t}-1)NB_{x})
	\end{equation}
	\begin{equation}
	(\overrightarrow{f}^{(i)}-\overrightarrow{B}^{(i)})_{I_{2}}=\overrightarrow{0},i=NB_{x},2NB_{x},...,NB_{t}\text{x}NB_{x}
	\end{equation}
	\item $\overrightarrow{\xi}_{ic}^{(i)}=\overrightarrow{0}$ 
	\begin{equation}
	(\overrightarrow{f}^{(i)}-\overrightarrow{F}^{(i)})_{I_{1}}=\overrightarrow{0},i=1,2,...,NB_{x}
	\end{equation}
\end{enumerate}

\subsubsection{Additional interface equations}
\begin{enumerate}
	\item Constraints for $C^{0}$ solutions i.e. $\overrightarrow{\xi}_{C^{0}}^{(i)}=\overrightarrow{0}$.
	\begin{itemize}
		\item Continuity along $x$ direction
		\begin{equation}
		\left\{ \begin{array}{c}
		\overrightarrow{f}^{(\kappa(1)+i-1)}\\
		\overrightarrow{f}^{(\kappa(2)+i-1)}\\
		...\\
		\overrightarrow{f}^{(\kappa(NB_{t})+i-1)}
		\end{array}\right\} _{I_{2}}=\left\{ \begin{array}{c}
		\overrightarrow{f}^{(\kappa(1)+i)}\\
		\overrightarrow{f}^{(\kappa(2)+i)}\\
		...\\
		\overrightarrow{f}^{(\kappa(NB_{t})+i)}
		\end{array}\right\} _{I_{4}},i=1,2,...,NB_{x}-1
		\end{equation}
		where $\kappa=[1,(1+NB_{x}),...,1+(NB_{t}-1)NB_{x}]^{T}$. 
		\item Continuity along $t$ direction
		\begin{equation}
		\left\{ \begin{array}{c}
		\overrightarrow{f}^{(\kappa(1)+(i-1)NB_{x})}\\
		\overrightarrow{f}^{(\kappa(2)+(i-1)NB_{x})}\\
		...\\
		\overrightarrow{f}^{(\kappa(NB_{x})+(i-1)NB_{x})}
		\end{array}\right\} _{I_{3}}=\left\{ \begin{array}{c}
		\overrightarrow{f}^{(\kappa(1)+iNB_{x})}\\
		\overrightarrow{f}^{(\kappa(2)+iNB_{x})}\\
		...\\
		\overrightarrow{f}^{(\kappa(NB_{x})+iNB_{x})}
		\end{array}\right\} _{I_{1}},i=1,2,...,NB_{t}-1
		\end{equation}
		where $\kappa=[1,2,...,NB_{x}]^{T}$. 
	\end{itemize}
	\item Constraints for $C^{1}$ solutions i.e. $\overrightarrow{\xi}_{C^{1}}^{(i)}=\overrightarrow{0}$.
	\begin{itemize}
		\item Smooth solutions along $x$ direction
		\begin{equation}
		\frac{\partial}{\partial x}\left\{ \begin{array}{c}
		\overrightarrow{f}^{(\kappa(1)+i-1)}\\
		\overrightarrow{f}^{(\kappa(2)+i-1)}\\
		...\\
		\overrightarrow{f}^{(\kappa(NB_{t})+i-1)}
		\end{array}\right\} _{I_{2}}=\frac{\partial}{\partial x}\left\{ \begin{array}{c}
		\overrightarrow{f}^{(\kappa(1)+i)}\\
		\overrightarrow{f}^{(\kappa(2)+i)}\\
		...\\
		\overrightarrow{f}^{(\kappa(NB_{t})+i)}
		\end{array}\right\} _{I_{4}},i=1,2,...,NB_{x}-1\label{eq:DPIELM_end}
		\end{equation}
		where $\kappa=[1,(1+NB_{x}),...,1+(NB_{t}-1)NB_{x}]^{T}$. 
	\end{itemize}
\end{enumerate}
Assembly of Eqns (\ref{eq:DPIELM_start} to \ref{eq:DPIELM_end})
leads to a system of linear equations which can be represented as
\begin{equation}
\boldsymbol{H}\overrightarrow{c}=\overrightarrow{K},\label{eq:Hc=00003DK-1}
\end{equation}
where $\overrightarrow{c}=[\overrightarrow{c}^{(1)},\overrightarrow{c}^{(2)},...,\overrightarrow{c}^{(N_{c})}]^{T}.$
The form of $\boldsymbol{H}$ and $\overrightarrow{K}$ depends on
the $\mathcal{L}$, $B$ and $F$. Finally $\overrightarrow{c}$ can
be found using pseudo inverse. It is to be noted that although we
have shown the formulation for 1D unsteady problems, no special adjustment
is needed to extend the formulation to higher dimensional problems.

This completes the mathematical formulation of DPIELM. The main steps
in its implementation are as follows:
\begin{enumerate}
	\item Divide the computational domain into uniformly distributed non overlapping
	cells and install a PIELM in each cell.
	\item Depending on the cell location, PDE and the initial and boundary conditions,
	find the expressions for $\overrightarrow{\xi}_{f}^{(i)}$, $\overrightarrow{\xi}_{bc}^{(i)}$
	and $\overrightarrow{\xi}_{ic}^{(i)}$ at each cell.
	\item Depending on the PDE, find the expressions for $\overrightarrow{\xi}_{C^{0}}^{(i)}$
	and $\overrightarrow{\xi}_{C^{1}}^{(i)}$ at each cell interface.
	\item Assemble these equations in the form of $\boldsymbol{H}\overrightarrow{c}=\overrightarrow{K},$
	where $\overrightarrow{c}=[\overrightarrow{c}^{(1)},\overrightarrow{c}^{(2)},...,\overrightarrow{c}^{(N_{c})}]^{T}.$
	\item Find the value of $\overrightarrow{c}$ using pseudo inverse.
\end{enumerate}
\section{\label{sec:7}Performance evaluation of DPIELM}

\begin{table}[H]
	\begin{centering}
		\begin{tabular}{|>{\centering}p{1.8cm}|>{\centering}p{5.5cm}|>{\centering}p{4.5cm}|}
			\hline 
			Test Case ID & Description & Architecture\tabularnewline
			\hline 
			\hline 
			TC-9 & 1D unsteady linear advection-diffusion & $[10,10,5,5,30]$\tabularnewline
			\hline 
			TC-10 & 2D unsteady linear advection-diffusion equation & $[20,20,50,3,3,3,30]$\tabularnewline
			\hline 
			TC-11 & Representation of 1D sharp and discontinuous functions  & $[50,5,5]$\tabularnewline
			\hline 
			TC-12 & Representation of a sharp peaked 2D Gaussian with DPIELM & $[15,15,5,5,15]$\tabularnewline
			\hline 
			TC-13 & Advection of a sharp peaked Gaussian & $[15,10,5,5,30]$\tabularnewline
			\hline 
			TC-14 & 1D steady advection-diffusion & $[20,5,20]$\tabularnewline
			\hline 
			TC-15 & Advection of a high frequency wave packet & $[15,10,5,5,30]$\tabularnewline
			\hline 
		\end{tabular}
		\par\end{centering}
	\caption{\label{tab:dpielm_details}Details of DPIELM architecture for the
		test cases. For 1D steady problems, architecture is given by $[NB_{x},nb_{x},N_{cell}^{*}]$,
		where $NB_{x}$ ,$nb_{x}$ and $N_{cell}^{*}$ refer to number of
		cells, number of points in the cell and size of hidden layer of the
		PIELM. Similarly, for 1D and 2D unsteady problems, it is given by
		$[NB_{x},NB_{t},nb_{x},nb_{t},N_{cell}^{*}]$ and $[NB_{x},NB_{y},NB_{t},nb_{x},nb_{y},nb_{t},N_{cell}^{*}]$
		respectively. }
	
\end{table}
In this section, we evaluate the performance of DPIELMs by testing
it on all the cases in which regular PIELM and PINN failed to perform.
The details of the architecture is given in Tab(\ref{tab:dpielm_details}).

\subsubsection*{Representation of functions with sharp gradient {[} TC-11, TC-12{]}}

\begin{figure}[H]
	\begin{centering}
		\includegraphics[scale=0.65]{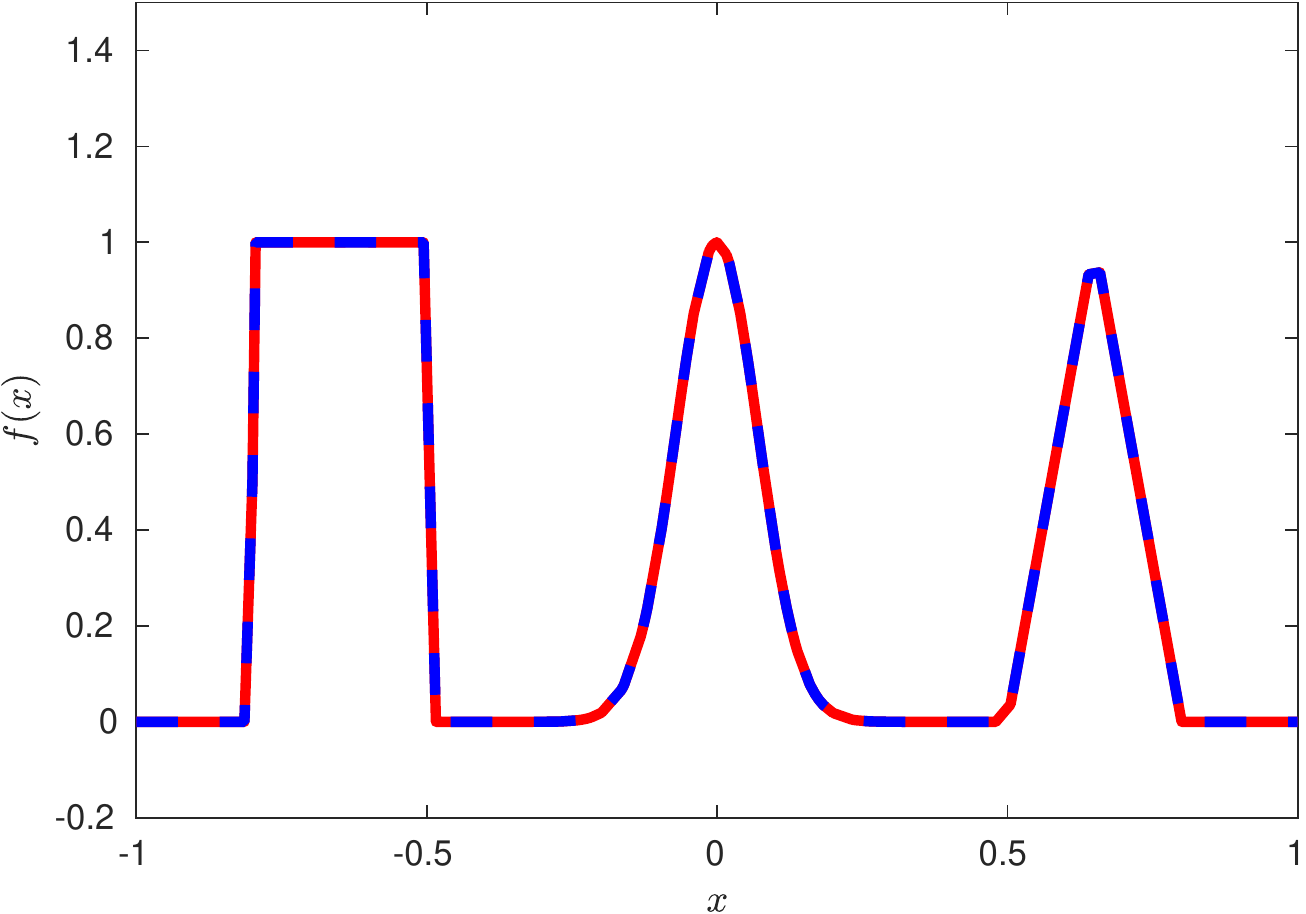}
		\par\end{centering}
	\caption{\label{fig:DPIELM_1D_rep}Representation of 1D non smooth functions
		with DPIELM. Red: DPIELM, Blue: Exact.}
\end{figure}
\begin{figure}[H]
	\begin{centering}
		\includegraphics[scale=0.7]{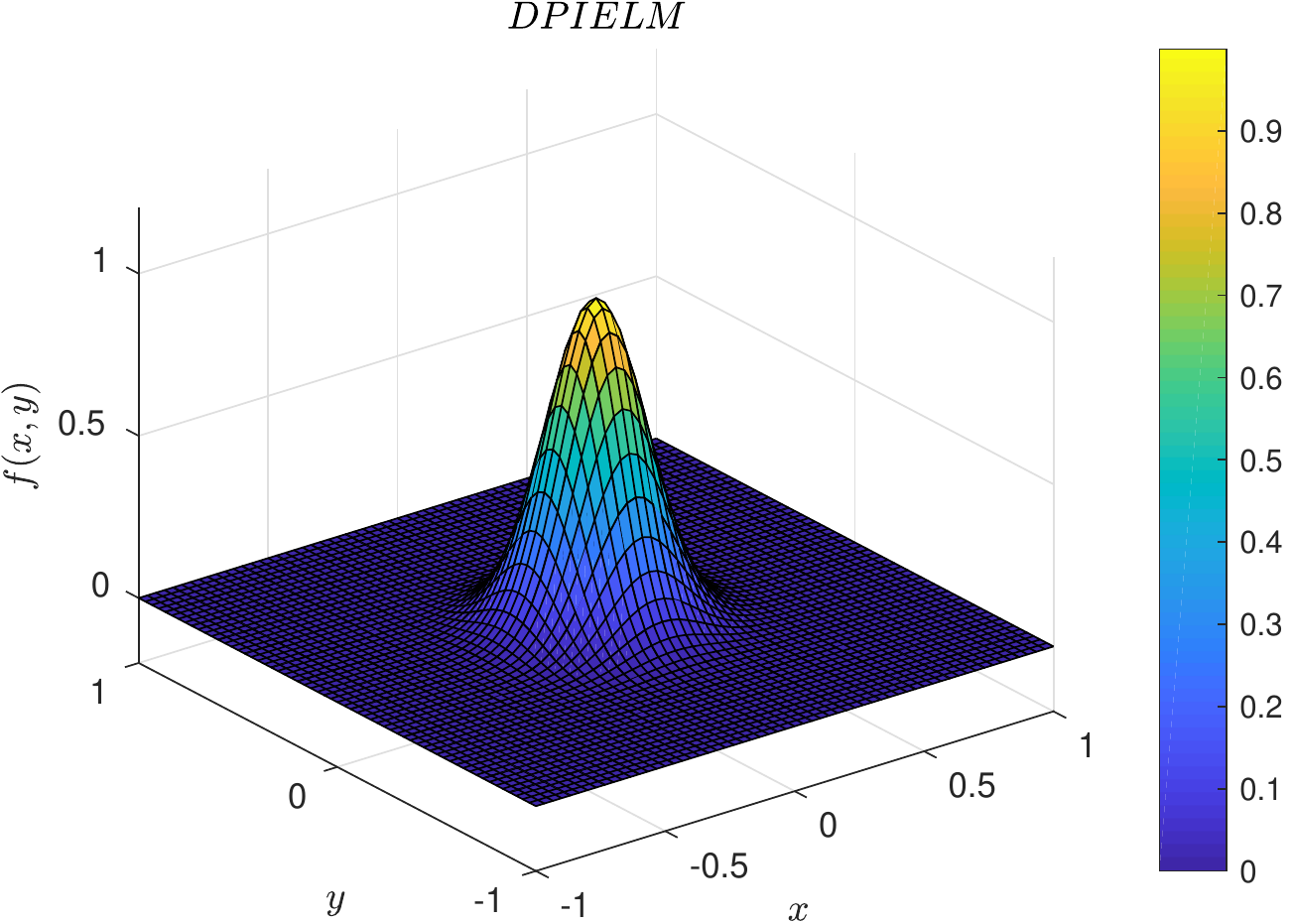}
		\par\end{centering}
	\caption{\label{fig:DPIELM_3D_rep}Representation of a sharp peaked 2D Gaussian
		with DPIELM}
\end{figure}
The mathematical formulation of DPIELM equations for 1D case are
as follows:
\begin{enumerate}
	\item $\text{\ensuremath{\overrightarrow{\xi}_{f}^{(i)}}=\ensuremath{\overrightarrow{0}}}$
	\begin{equation}
	\varphi(\boldsymbol{X_{f}}^{(i)}\boldsymbol{W}^{(i)^{T}})\text{\ensuremath{\odot}}\overrightarrow{c}^{(i)}=f(\overrightarrow{x}_{f}^{(i)}),i=1,2,...,NB_{x}
	\end{equation}
	\item $\overrightarrow{\xi}_{bc}^{(i)}=\overrightarrow{0}$
	\begin{equation}
	\left\{ \begin{array}{c}
	\varphi(\boldsymbol{X_{bc,I_{1}}}^{(1)}\boldsymbol{W}^{(1)^{T}})\overrightarrow{c}^{(1)}\\
	\varphi(\boldsymbol{X_{bc,I_{2}}}^{(NB_{x})}\boldsymbol{W}^{(NB_{x})^{T}})\overrightarrow{c}^{(NB_{x})}
	\end{array}\right\} =\left\{ \begin{array}{c}
	f(\overrightarrow{x}_{bc,I_{1}}^{(1)})\\
	f(\overrightarrow{x}_{bc,I_{2}}^{(NB_{x})})
	\end{array}\right\} 
	\end{equation}
	where $I_{1},I_{2}$ refer to left and right cell interfaces respectively.
	\item $\overrightarrow{\xi}_{C^{0}}^{(i)}=\overrightarrow{0}$
	\begin{equation}
	\varphi(\boldsymbol{X_{bc,I_{2}}}^{(i)}\boldsymbol{W}^{(i)^{T}})\overrightarrow{c}^{(i)}=\varphi(\boldsymbol{X_{bc,I_{1}}}^{(i+1)}\boldsymbol{W}^{(i+1)^{T}})\overrightarrow{c}^{(i+1)},i=1,2,...,NB_{x}-1
	\end{equation}
\end{enumerate}
The equations for the 3D case can be written in a similar fashion.
The exact and DPIELM solutions for these cases are shown in Figs (\ref{fig:DPIELM_1D_rep})
and (\ref{fig:DPIELM_3D_rep}) respectively. 

\subsubsection*{1D steady advection-diffusion equation with low value of diffusion
	constant {[}TC-14{]}}

\begin{figure}[H]
	\begin{centering}
		\includegraphics[scale=0.75]{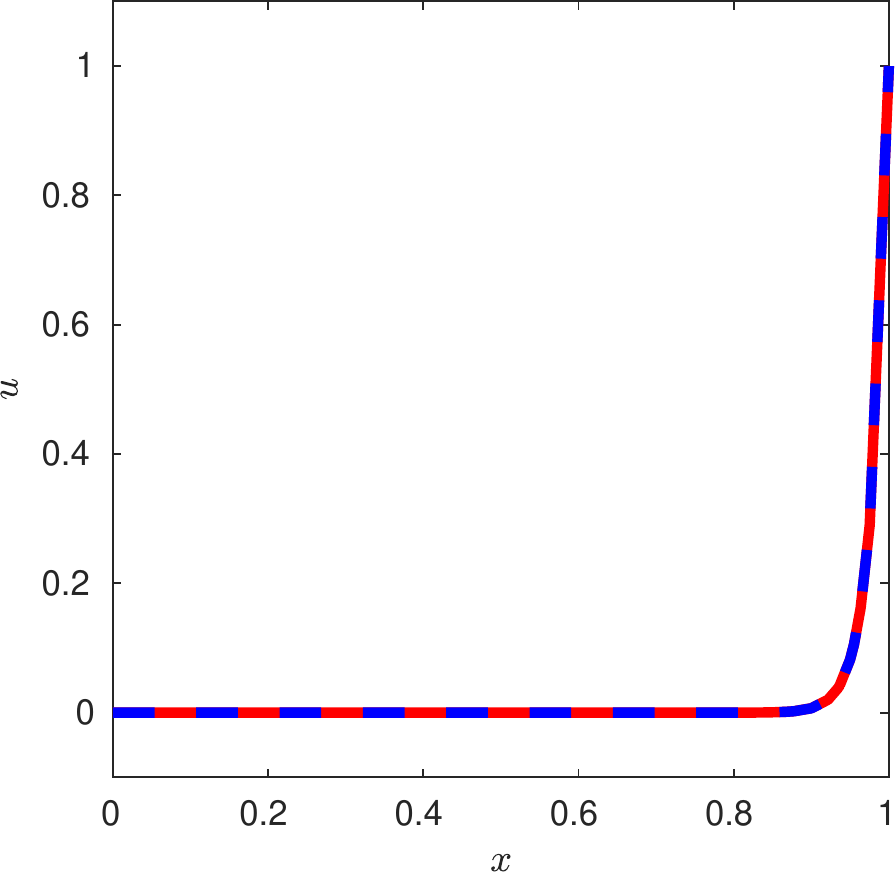}
		\par\end{centering}
	\caption{\label{fig:DPIELM_SAD_02}Exact and DPIELM solution of unsteady 1D
		convection diffusion. Red: DPIELM, Blue: exact}
\end{figure}
The DPIELM equations are as follows:
\begin{enumerate}
	\item $\text{\ensuremath{\overrightarrow{\xi}_{f}^{(i)}}=\ensuremath{\overrightarrow{0}}}$
	\begin{equation}
	\varphi'(\boldsymbol{X_{f}}^{(i)}\boldsymbol{W}^{(i)^{T}})\text{\ensuremath{\odot}}\overrightarrow{c}^{(i)}\text{\ensuremath{\odot}}\overrightarrow{m}-\nu\varphi''(\boldsymbol{X_{f}}^{(i)}\boldsymbol{W}^{(i)^{T}})\text{\ensuremath{\odot}}\overrightarrow{c}\text{\ensuremath{\odot}}\overrightarrow{m}\text{\ensuremath{\odot}}\overrightarrow{m}=\overrightarrow{0},i=1,2,...,NB_{x}
	\end{equation}
	\item $\overrightarrow{\xi}_{bc}^{(i)}=\overrightarrow{0}$
	\begin{equation}
	\left\{ \begin{array}{c}
	\varphi(\boldsymbol{X_{bc,I_{1}}}^{(1)}\boldsymbol{W}^{(1)^{T}})\overrightarrow{c}^{(1)}\\
	\varphi(\boldsymbol{X_{bc,I_{2}}}^{(NB_{x})}\boldsymbol{W}^{(NB_{x})^{T}})\overrightarrow{c}^{(NB_{x})}
	\end{array}\right\} =\left\{ \begin{array}{c}
	B(\overrightarrow{x}_{bc,I_{1}}^{(1)})\\
	B(\overrightarrow{x}_{bc,I_{2}}^{(NB_{x})})
	\end{array}\right\} 
	\end{equation}
	\item $\overrightarrow{\xi}_{C^{0}}^{(i)}=\overrightarrow{0}$
	\begin{equation}
	\varphi(\boldsymbol{X_{bc,I_{2}}}^{(i)}\boldsymbol{W}^{(i)^{T}})\overrightarrow{c}^{(i)}=\varphi(\boldsymbol{X_{bc,I_{1}}}^{(i+1)}\boldsymbol{W}^{(i+1)^{T}})\overrightarrow{c}^{(i+1)},i=1,2,...,NB_{x}-1
	\end{equation}
	\item $\overrightarrow{\xi}_{C^{1}}^{(i)}=\overrightarrow{0}$
	\begin{equation}
	\varphi'(\boldsymbol{X_{bc,I_{2}}}^{(i)}\boldsymbol{W}^{(i)^{T}})\overrightarrow{c}^{(i)}\text{\ensuremath{\odot}}\overrightarrow{m}^{(i)}=\varphi'(\boldsymbol{X_{bc,I_{1}}}^{(i+1)}\boldsymbol{W}^{(i+1)^{T}})\overrightarrow{c}^{(i+1)}\text{\ensuremath{\odot}}\overrightarrow{m}^{(i+1)},i=1,2,...,NB_{x}-1
	\end{equation}
\end{enumerate}
The results of the exact and DPIELM solution is given in Fig(\ref{fig:DPIELM_SAD_02}).

\subsection{1D unsteady advection of a sharp peaked Gaussian and a high frequency
	wavelet and {[} TC-13, TC-15 {]}}

\begin{figure}[H]
	\begin{centering}
		\includegraphics[scale=0.6]{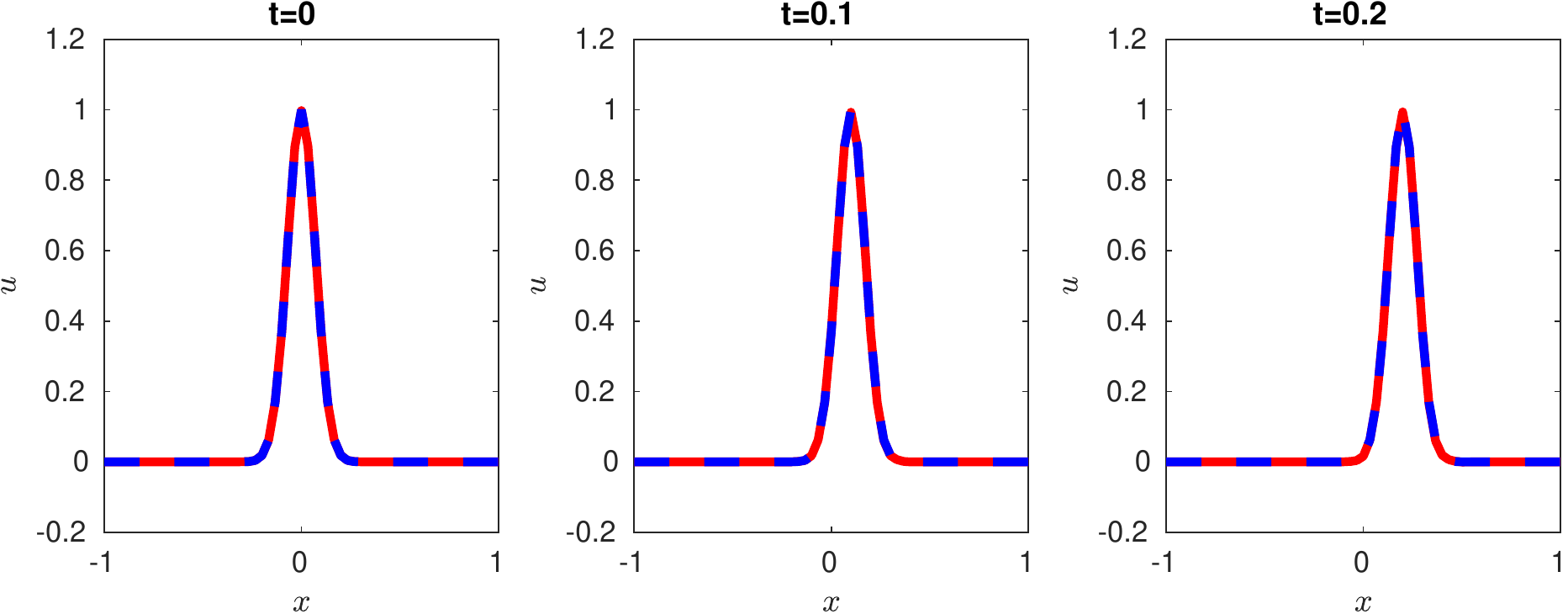}
		\par\end{centering}
	\caption{\label{fig:DPIELM_1D_bell_advect}Exact and DPIELM solution advection
		of a sharp peaked Gaussian with DPIELM. Red: DPIELM, Blue: Exact.}
\end{figure}
\begin{figure}[H]
	\begin{centering}
		\includegraphics[scale=0.5]{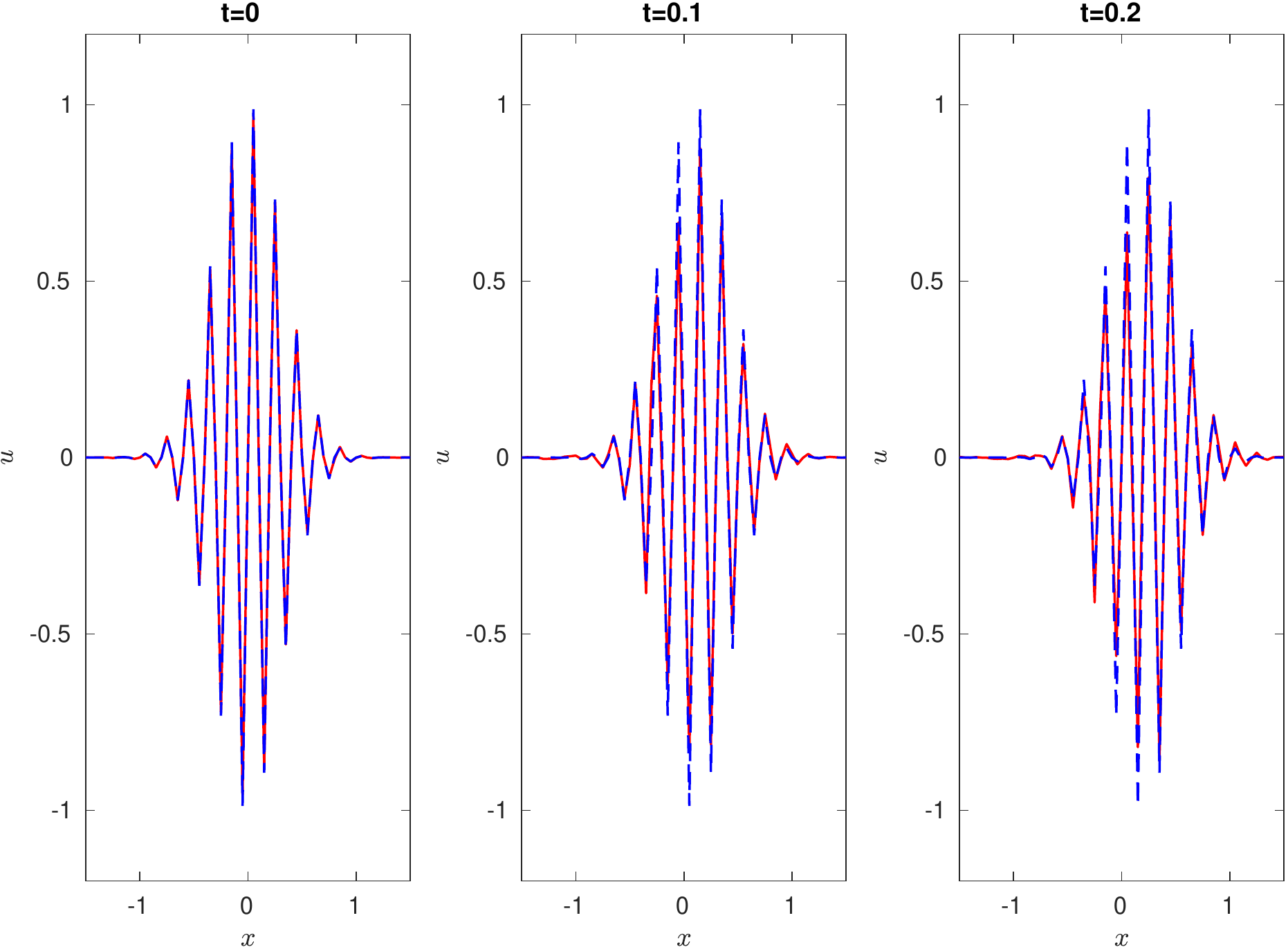}
		\par\end{centering}
	\caption{\label{fig:DPIELM_wavelet}Exact and DPIELM solution of pure advection
		of a high frequency wave packet. Red: DPIELM, Blue: exact}
\end{figure}
The DPIELM equation to be solved are as follows: 
\begin{enumerate}
	\item $\text{\ensuremath{\overrightarrow{\xi}_{f}^{(i)}}=\ensuremath{\overrightarrow{0}} }$
	\begin{equation}
	\varphi'(\boldsymbol{X_{f}}^{(i)}\boldsymbol{W}^{(i)^{T}})\text{\ensuremath{\odot}}\overrightarrow{c}^{(i)}\text{\ensuremath{\odot}}(\overrightarrow{m}^{(i)}+\overrightarrow{n}^{(i)})=\overrightarrow{0},i=1,2,...,N_{c}\label{eq:DPIELM_adv}
	\end{equation}
	
	\item $\overrightarrow{\xi}_{bc}^{(i)}=\overrightarrow{0}$ 
	\begin{equation}
	\varphi(\boldsymbol{X_{bc,I_{4}}}^{(i)}\boldsymbol{W}^{(i)^{T}})\overrightarrow{c}^{(i)}=\overrightarrow{0},i=1,(1+NB_{x}),...,(1+(NB_{t}-1)NB_{x})
	\end{equation}
	\begin{equation}
	\varphi(\boldsymbol{X_{bc,I_{2}}}^{(i)}\boldsymbol{W}^{(i)^{T}})\overrightarrow{c}^{(i)}=\overrightarrow{0},i=NB_{x},2NB_{x},...,NB_{t}\text{x}NB_{x}
	\end{equation}
	\item $\overrightarrow{\xi}_{ic}^{(i)}=\overrightarrow{0}$ 
	\begin{equation}
	\varphi(\boldsymbol{X_{bc,I_{1}}}^{(i)}\boldsymbol{W}^{(i)^{T}})\overrightarrow{c}^{(i)}=\overrightarrow{F}(\overrightarrow{x}_{bc,I_{1}}^{(i)}),i=1,2,...,NB_{x}
	\end{equation}
	\item $\overrightarrow{\xi}_{C^{0}}^{(i)}=\overrightarrow{0}$
	\begin{itemize}
		\item Continuity along $x$ direction
		\begin{equation}
		\begin{array}{c}
		\left\{ \begin{array}{c}
		\varphi(\boldsymbol{X_{bc,I_{2}}}^{(j(1,i)-1)}\boldsymbol{W}^{(j(1,i)-1)^{T}})\overrightarrow{c}^{(j(1,i)-1)}\\
		\varphi(\boldsymbol{X_{bc,I_{2}}}^{(j(2,i)-1)}\boldsymbol{W}^{(j(2,i)-1)^{T}})\overrightarrow{c}^{(j(2,i)-1)}\\
		...\\
		\varphi(\boldsymbol{X_{bc,I_{2}}}^{(j(NB_{t},i)-1)}\boldsymbol{W}^{(j(NB_{t},i)-1)^{T}})\overrightarrow{c}^{(j(NB_{t},i)-1)}
		\end{array}\right\} =\\
		\left\{ \begin{array}{c}
		\varphi(\boldsymbol{X_{bc,I_{4}}}^{(j(1,i))}\boldsymbol{W}^{(j(1,i))^{T}})\overrightarrow{c}^{(j(1,i))}\\
		\varphi(\boldsymbol{X_{bc,I_{4}}}^{(j(2,i))}\boldsymbol{W}^{(j(2,i))^{T}})\overrightarrow{c}^{(j(2,i))}\\
		...\\
		\varphi(\boldsymbol{X_{bc,I_{I_{4}}}}^{(j(NB_{t},i))}\boldsymbol{W}^{(j(NB_{t},i))^{T}})\overrightarrow{c}^{(j(NB_{t},i))}
		\end{array}\right\} 
		\end{array}
		\end{equation}
		where $\rho=1,2...,NB_{t}$, $j(\rho,i)=\kappa(\rho)+i,$ $i=1,2,...,NB_{x}-1$
		and $\kappa=[1,(1+NB_{x}),...,1+(NB_{t}-1)NB_{x}]^{T}$. 
		\item Continuity along $t$ direction
		\begin{equation}
		\begin{array}{c}
		\left\{ \begin{array}{c}
		\varphi(\boldsymbol{X_{bc,I_{3}}}^{(j(1,i-1)}\boldsymbol{W}^{(j(1,i-1)^{T}})\overrightarrow{c}^{(j(1,i-1)}\\
		\varphi(\boldsymbol{X_{bc,I_{3}}}^{(j(2,i-1)}\boldsymbol{W}^{(j(2,i-1)^{T}})\overrightarrow{c}^{(j(2,i-1)}\\
		...\\
		\varphi(\boldsymbol{X_{bc,I_{3}}}^{(j(NB_{t},i-1)}\boldsymbol{W}^{(j(NB_{t},i-1)^{T}})\overrightarrow{c}^{(j(NB_{t},i-1)}
		\end{array}\right\} =\\
		\left\{ \begin{array}{c}
		\varphi(\boldsymbol{X_{bc,I_{1}}}^{(j(1,i))}\boldsymbol{W}^{(j(1,i))^{T}})\overrightarrow{c}^{(j(1,i))}\\
		\varphi(\boldsymbol{X_{bc,1}}^{(j(2,i))}\boldsymbol{W}^{(j(2,i))^{T}})\overrightarrow{c}^{(j(2,i))}\\
		...\\
		\varphi(\boldsymbol{X_{bc,I_{I_{1}}}}^{(j(NB_{t},i))}\boldsymbol{W}^{(j(NB_{t},i))^{T}})\overrightarrow{c}^{(j(NB_{t},i))}
		\end{array}\right\} 
		\end{array}
		\end{equation}
		where $\rho=1,2...,NB_{x}$, $j(\rho,i)=\kappa(\rho)+iNB_{x},$ $i=1,2,...,NB_{t}-1$
		and $\kappa=[1,2,...,NB_{x}]^{T}$. 
	\end{itemize}
\end{enumerate}
The results for the TC-13 and TC-15 are given in Fig (\ref{fig:DPIELM_1D_bell_advect})
and Fig (\ref{fig:DPIELM_wavelet}) respectively. 

\subsection{\label{subsec:FARHAT_DPIELM}1D and 2D unsteady advection-diffusion
	equations {[} TC-9, TC-10{]}}

\begin{figure}[H]
	\begin{centering}
		\includegraphics[scale=0.75]{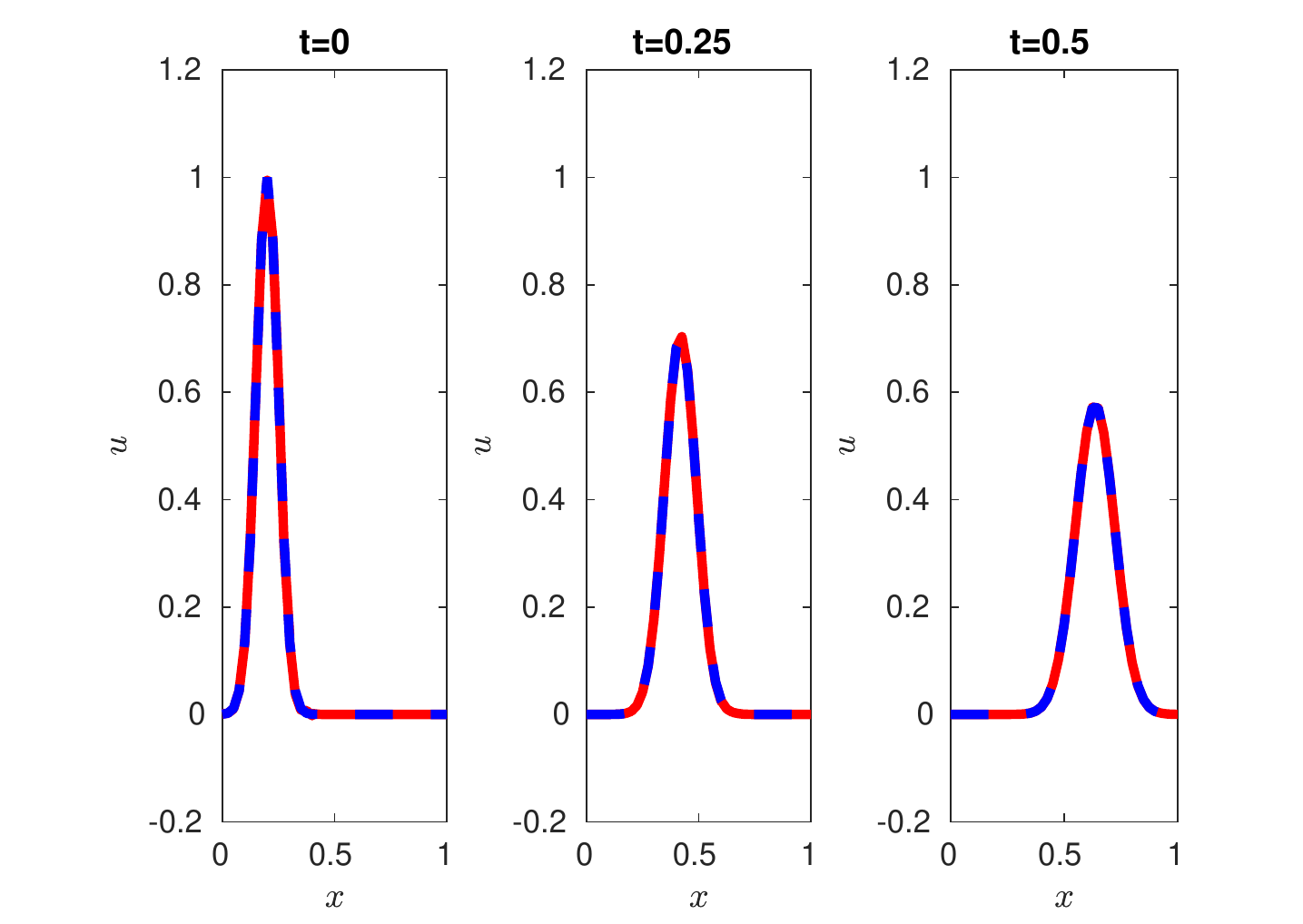}
		\par\end{centering}
	\caption{\label{fig:DPIELM_1D_CD}Exact and DPIELM solution of unsteady 1D
		convection diffusion. Red: DPIELM, Blue: exact}
\end{figure}
\begin{figure}[H]
	\includegraphics[scale=0.4]{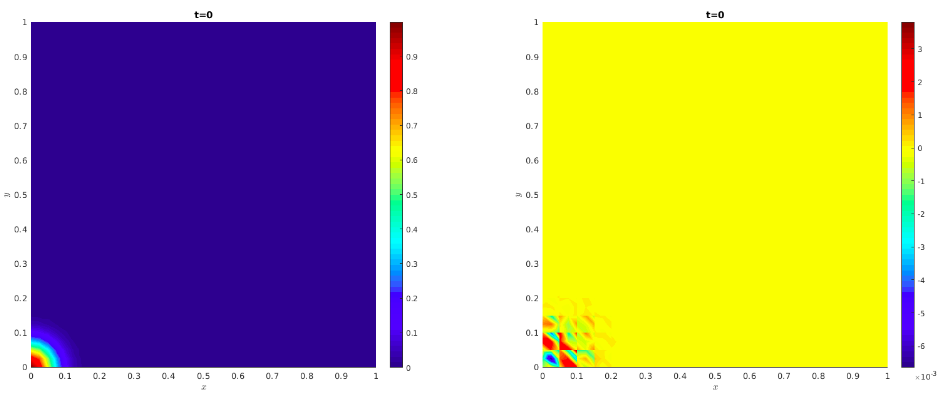}
	
	\caption{\label{fig:dpielm_t0}DPIELM solution and error for unsteady 2D convection
		diffusion at $t=0$ }
\end{figure}
\begin{figure}[H]
	\includegraphics[scale=0.4]{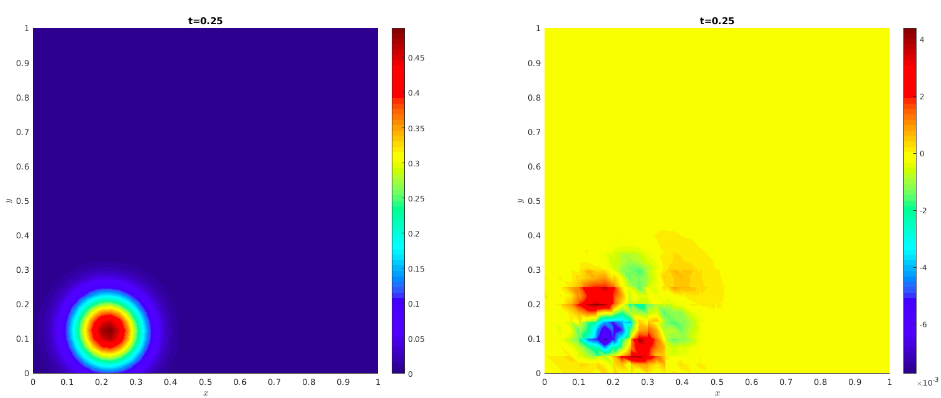}
	
	\caption{DPIELM solution and error for unsteady 2D convection diffusion at
		$t=0.25$ }
\end{figure}
\begin{figure}[H]
	\includegraphics[scale=0.4]{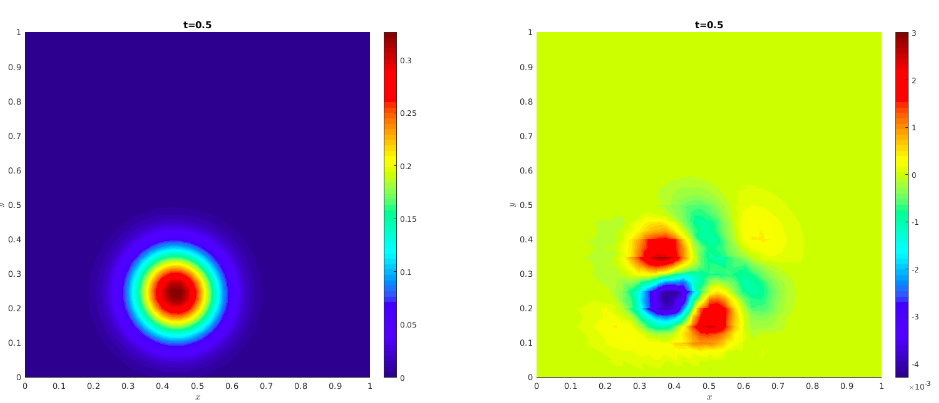}
	
	\caption{\label{fig:dpielm_tf}DPIELM solution and error for unsteady 2D convection
		diffusion at $t=0.5$ }
\end{figure}
In this section, we present the DPIELM equation for the 1D case. The
equations for the 2D case can be formulated in a similar fashion.
The equations to be solved for 1D unsteady advection-diffusion equation
are as follows:
\begin{enumerate}
	\item $\text{\ensuremath{\overrightarrow{\xi}_{f}^{(i)}}=\ensuremath{\overrightarrow{0}} }$
	\begin{equation}
	\varphi'(\boldsymbol{X_{f}}^{(i)}\boldsymbol{W}^{(i)^{T}})\text{\ensuremath{\odot}}\overrightarrow{c}^{(i)}\text{\ensuremath{\odot}}(\overrightarrow{n}^{(i)}+a\overrightarrow{m}^{(i)}-\nu\overrightarrow{m}^{(i)}\text{\ensuremath{\odot}}\overrightarrow{m}^{(i)})=\overrightarrow{0},i=1,2,...,N_{c}
	\end{equation}
	\item $\overrightarrow{\xi}_{C^{1}}^{(i)}=\overrightarrow{0}$ 
	\begin{equation}
	\begin{array}{c}
	\left\{ \begin{array}{c}
	\varphi'(\boldsymbol{X_{bc,I_{2}}}^{(j(1,i)-1)}\boldsymbol{W}^{(j(1,i)-1)^{T}})\overrightarrow{c}^{(j(1,i)-1)}\ensuremath{\odot\overrightarrow{m}^{(j(1,i)-1)}}\\
	\varphi'(\boldsymbol{X_{bc,I_{2}}}^{(j(2,i)-1)}\boldsymbol{W}^{(j(2,i)-1)^{T}})\overrightarrow{c}^{(j(2,i)-1)}\ensuremath{\odot\overrightarrow{m}^{(j(2,i)-1)}}\\
	...\\
	\varphi'(\boldsymbol{X_{bc,I_{2}}}^{(j(NB_{t},i)-1)}\boldsymbol{W}^{(j(NB_{t},i)-1)^{T}})\overrightarrow{c}^{(j(NB_{t},i)-1)}\odot\overrightarrow{m}^{(j(NB_{t},i)-1)}
	\end{array}\right\} =\\
	\left\{ \begin{array}{c}
	\varphi'(\boldsymbol{X_{bc,I_{4}}}^{(j(1,i))}\boldsymbol{W}^{(j(1,i))^{T}})\overrightarrow{c}^{(j(1,i))}\odot\overrightarrow{m}^{(j(1,i))}\\
	\varphi'(\boldsymbol{X_{bc,I_{4}}}^{(j(2,i))}\boldsymbol{W}^{(j(2,i))^{T}})\overrightarrow{c}^{(j(2,i))}\odot\overrightarrow{m}^{(j(2,i))}\\
	...\\
	\varphi'(\boldsymbol{X_{bc,I_{I_{4}}}}^{(j(NB_{t},i))}\boldsymbol{W}^{(j(NB_{t},i))^{T}})\overrightarrow{c}^{(j(NB_{t},i))}\odot\overrightarrow{m}^{(j(NB_{t},i))}
	\end{array}\right\} 
	\end{array}
	\end{equation}
	where $\rho=1,2...,NB_{t}$, $j(\rho,i)=\kappa(\rho)+i,$ $i=1,2,...,NB_{x}-1$
	and $\kappa=[1,(1+NB_{x}),...,1+(NB_{t}-1)NB_{x}]^{T}$. 
\end{enumerate}
Rest of the equations i.e. $\overrightarrow{\xi}_{bc}^{(i)}=\overrightarrow{0}$,
$\overrightarrow{\xi}_{ic}^{(i)}=\overrightarrow{0}$ and $\overrightarrow{\xi}_{C^{0}}^{(i)}=\overrightarrow{0}$
are same as that used in 1D unsteady linear advection. The results
for the 1D and 2D cases are given in Fig (\ref{fig:DPIELM_1D_CD})
and Fig (\ref{fig:dpielm_t0} to \ref{fig:dpielm_tf}) respectively. 

This brings us to the end of our numerical experiments. We close this
section by highlighting the key points which are as follows:
\begin{enumerate}
	\item The process of partitioning of the whole computational domain into
	multiple cells simplifies the representation of the complicated function
	( and thus PDE ) in the individual cells. As a result, local PIELMs
	are able to capture not just the functions with sharp gradients, but
	also discontinuous functions ( TC-11 ).
	\item To our knowledge, this is the first demonstration of capability of
	ELM based algorithms to solve 2D unsteady PDEs and produce results
	comparable to sophisticated numerical methods. (TC-10)
	\item The distributed version of PIELM exhibits better representation ability
	than a deep PINN (TC-15).
\end{enumerate}

\section{\label{sec:8}Conclusion and future work}
We have presented in this paper PIELM -{}- an efficient method to
utilize physics informed neural networks to solve stationary and time
dependent linear PDEs. As PIELM inherits the unique qualities of its
parent algorithms (PINN and ELM), it works very well on complex geometries,
respects the inherent physics of the PDEs and is extremely fast as
well. This leads to several advantages over existing conventional
numerical methods; PIELM reduces numerical artefacts such as false
diffusion as well can handle complex geometries in a meshfree approach.
PIELM also reduces, to a certain extent, the arbitrariness of the
number of neurons in typical deep PINNs. Our numerical tests also
confirm that, for a fixed problem, our minimal PIELM is more accurate
than prior deep NN results \cite{BERG ET AL} while being faster.

We have also presented in this paper the limitations in representing
complex functions using a single PIELM or PINN for the whole domain.
For practical problems using PINNs can lead to very deep networks
with the concomitant training problems and efficiency issues. Our
proposed solution is to use a distributed PIELM (DPIELM) which uses
different representations in different portions of the domain while
imposing some continuity and differentiability constraints. The resultant
DPIELM easily captures profiles that PINNs have difficulty with. Further,
on time-dependent problems, DPIELM gives results that are comparable
to sophisticated conventional numerical techniques as seen in Section
\ref{subsec:FARHAT_DPIELM}.

We believe that the method, as formulated, is already very powerful
for linear PDEs with constant or space-varying coefficients. Two primary
areas of development remain, in our opinion. Our preliminary tests
show that, for linear PDEs, unlike deep PINNs \cite{RAISSI PINN},
our method may actually be competitive with conventional techniques
in terms of speed and accuracy. However, a firm conclusion on this
cannot be reached until a more thorough and fair study is done. Theoretical
and numerical evidence for this efficacy is the first area which deserves
attention, as the number of practical applications (such as heat conduction,
etc) where numerical methods for linear equations are a staple is
large. Having an efficient neural network framework for such problems
can be tremendously beneficial to practitioners.

Even more importantly, PIELM\textquoteright s efficacy is thanks to
the linear nature of the final problem. We have, therefore, limited
our present study to linear problems. Extending this method to nonlinear
equations is the obvious next frontier. This may be approached in
one of two ways. The first approach would be to use the PIELM structure
as is and then solve the resultant, non-convex optimization problem.
Another approach would be to linearize the equation around the current
time step to predict a future time step. This would be the equivalent
of a linearized, explicit time-stepping method in conventional time
marching techniques. We are currently investigating these approaches
and will report progress in future publications.

\bibliographystyle{unsrt}  


\end{document}